\documentclass[10pt,journal,compsoc]{IEEEtran}
\usepackage{tabularx,booktabs}
\newcolumntype{C}{>{\centering\arraybackslash}X} 
\usepackage{xcolor}
\usepackage{wrapfig,lipsum,booktabs}
\usepackage[super]{nth}
\usepackage{float}
\usepackage{hyperref}
\hypersetup{
    filecolor=magenta,      
    urlcolor=cyan,
}

\usepackage[bottom]{footmisc}
\usepackage[justification=raggedright,singlelinecheck=true,tableposition=top]{caption}
\usepackage{commath}

\usepackage{graphicx}

\ifCLASSINFOpdf
\else
\fi

\usepackage{multirow}

\begin{document}
%
\title{Forecasting Action through Contact Representations from First Person Video}

\author{Eadom Dessalene\textbf{*}\thanks{$\bullet$ \:The authors are with the department of Computer Science, University of Maryland, College Park, MD, 20742}\thanks{$\bullet$ \:*Equal contribution},
        Chinmaya Devaraj\textbf{*}, Michael Maynord\textbf{*}, Cornelia Ferm\"uller,
        and Yiannis Aloimonos}


\IEEEtitleabstractindextext{%
\begin{abstract}



Human actions involving hand manipulations are structured according to the making and breaking of hand-object contact, and human visual understanding of action is reliant on anticipation of contact as is demonstrated by pioneering work in cognitive science. Taking inspiration from this, we introduce representations and models centered on contact, which we then use in action prediction and anticipation. We annotate a subset of the EPIC Kitchens dataset to include time-to-contact between hands and objects, as well as segmentations of hands and objects. Using these annotations we train the \textit{Anticipation Module}, a module producing \textit{Contact Anticipation Maps} and \textit{Next Active Object Segmentations} - novel low-level representations providing temporal and spatial characteristics of anticipated near future action. On top of the Anticipation Module we apply \textit{Egocentric Object Manipulation Graphs} (Ego-OMG), a framework for action anticipation and prediction. Ego-OMG models longer term temporal semantic relations through the use of a graph modeling transitions between contact delineated action states. Use of the Anticipation Module within Ego-OMG produces state-of-the-art results, achieving \nth{1} and \nth{2} place on the unseen and seen test sets, respectively, of the EPIC Kitchens Action Anticipation Challenge, and achieving state-of-the-art results on the tasks of action anticipation and action prediction over EPIC Kitchens. We perform ablation studies over characteristics of the Anticipation Module to evaluate their utility.

\end{abstract}

}

\maketitle

\IEEEdisplaynontitleabstractindextext

\IEEEpeerreviewmaketitle

\section{Introduction}

Understanding and anticipating others' actions is a necessary capability for fluid human interaction and collaboration. Without this capability collaboration involves excessive wait times as we wait for others' actions to complete. Responding earlier to others' actions reduces physical load, cognitive load, and the completion time of the task \cite{sexton2018anticipation, vesper2014support}.



\begin{figure}[h]
\includegraphics[width=9.2cm]{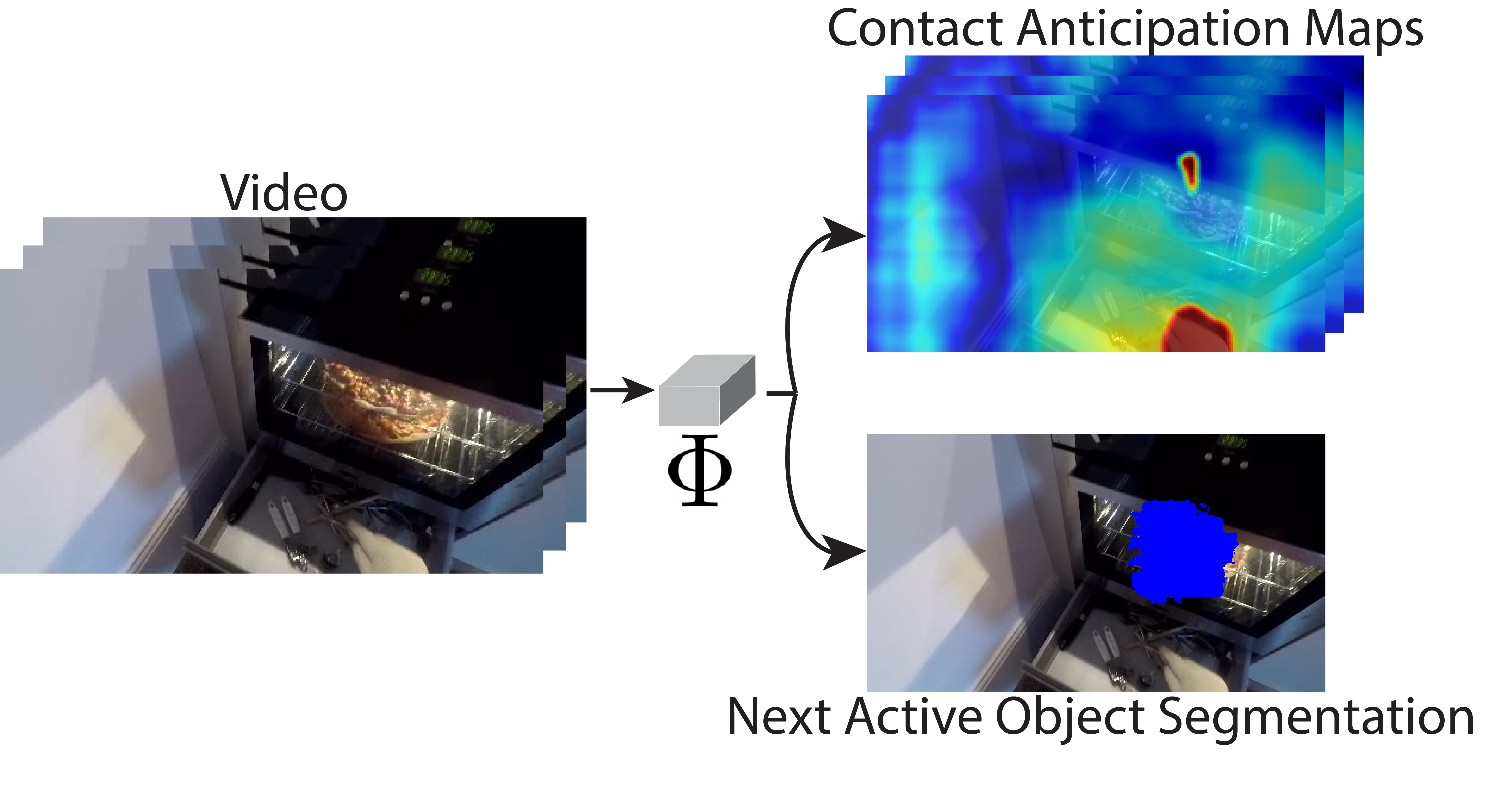}
\caption{Illustration of representations involved in the Anticipation Module $\Phi$ where the scenario depicts a person reaching into the oven. RGB video feeds into the Anticipation Module which produces Contact Anticipation Maps and a localization of the Next Active Object Segmentation. Visualization colors in the Contact Anticipation Maps vary from blue (large time to contact) to red (pixels belonging to hands or objects in contact).}
\label{fig:figure1}
\end{figure}

In our work on action understanding, we leverage first person - or egocentric - perspective, rather than the third person perspective more common in action datasets. There are a few reasons for this: 1) the egocentric perspective provides a less occluded view of the hands and the action being performed, 2) this view contains cues of intentionality – e.g., we tend to look towards the destination or focus of our actions,
3) as head mounted displays, including augmented reality headsets, become more common, egocentric data is becoming more readily available and methods involving the egocentric perspective more relevant (and, robots are able to leverage egocentric data from human worn sensors). In this paper we work with the egocentric datasets EPIC Kitchens \cite{damen2018scaling} and EGTEA \cite{li2018eye}.






Human interaction with the environment is largely performed through hand manipulations of objects. Each manipulation involves the making and breaking of hand object contact as a defining characteristic. Possible benefits of contact include better: 1) determining the class of action being performed, 2) delineating action boundaries, and 3) projection into the near future of action. As such, we structure our representations around contact with objects.




We define two classes of object, aligning with two different times of interest: the present, and the near future. Previous works \cite{pirsiavash2012detecting} have defined an \textit{Active Object} as an object currently involved in a given interaction. In this work, we define the \textit{Active Object} of a hand as the object presently in contact with the hand, and we define the \textit{Next Active Object} as the object which will next come into contact with that hand. In seeking to model future action we produce predictions for the Next Active Object.


Understanding which objects are Active Objects involves understanding hand-object contact. Understanding Next Active Objects involves predicting future hand-object contact. There is evidence from the cognitive science literature that modeling of contact plays a central role in human visual understanding of action \cite{tresilian1995perceptual, zago2009visuo}. As such, we center our models around contact.


We introduce components and representations useful for understanding contact. The \textit{Anticipation Module} contains two networks: the \textit{Contact Anticipation Network}, and the \textit{Next Active Object Network}. The \textit{Contact Anticipation Network} produces a representation termed the \textit{Contact Anticipation Map}, and the \textit{Next Active Object Network} produces a \textit{Next Active Object Segmentation}. See Figure \ref{fig:figure1} for an illustration of the representations produced by the Anticipation Module.



Pioneering works in cognitive science (e.g., \cite{jeannerod1984timing}) indicate that the velocity profiles of point-to-point hand movements follow a bell shaped distribution. We verify the presence of this bell shaped distribution with experiments: See Section 3 of the Supplementary Material for an illustration. During the onset of hand motion, the hand gradually accelerates, and as the hand approaches contact it rapidly decelerates. This is a motion cue relevant to action on which humans rely when understanding each other's actions \cite{stadler2012movement}. This shows that in hand reaching there is structure in the relations between the position of the hand, the position of the object, and the velocity of the hand. This low-level cue is of central relevance to action understanding, particularly anticipation of near future action characteristics, and we model it through the Contact Anticipation Maps.


Contact Anticipation Maps are a hand-centric representation, providing a pixel-wise estimation of potential time-to-contact between the hand and pixels in the scene. Pixels belonging to the hands of the actor and Active Object(s) are represented with time-to-contact values of 0. See Figure \ref{fig:figure1} for illustration.

The Contact Anticipation Network produces Contact Anticipation Maps in a low level fashion, without utilizing components or representations critically dependent upon accurate performance of object detectors, hand trackers, the category of the object being acted upon, or classification of the action being performed. This low-level approach to anticipating the next active object is a less brittle approach than approaches critically dependent on the performance of object detectors and hand trackers.


We feed a history of Contact Anticipation Maps in parallel with a stack of RGB frames to a network whose purpose is to localize the Next Active Object - the Next Active Object Network. The Next Active Object Network produces a representation localizing the likely Next Active Object - the Next Active Object Segmentation. In combination the Contact Anticipation Network and the Next Active Object Network provide a prediction for \textit{where} in the scene the Next Active Object will be, and \textit{when} contact with that object will be established.




The Next Active Object Segmentation is useful in understanding the type of interaction which will take place. Segmentation provides cues such as size, shape, and distance from the person, as well as providing a specific localization over which object classification can be run, providing an object category. 


To produce data with which to train the Anticipation Module we augment a portion of the EPIC Kitchens dataset with annotations of hands and objects, and the times at which hand / object contact occurs. This allows us to construct, at each frame prior to contact, a pixel level labeling of the hand, the Next Active Object, and the time remaining until that object and the hand come into contact. EPIC Kitchens provides RGB data, and includes no depth data - and while hand trajectories are best represented in 3 dimensions, 2 dimensional projections still provide ample trajectory information.



In our work on action understanding we approach two related tasks: action prediction, and action anticipation. Action prediction is the task of recognizing an action given only a partial observation of an ongoing action. Action anticipation is the task of anticipating the category of a near future action before its start. The representations produced by the Anticipation Module are of utility to the tasks of action prediction and anticipation, and we evaluate the anticipation module w.r.t. performance on these tasks.

Not only are the short range action characteristics provided by the Anticipation Module relevant to these tasks, but longer-range activity structure is relevant as well. For the modeling of longer range context and relations, methods beyond the Anticipation Module are needed. We extend the temporal window of activity modeling with Egocentric Object Manipulation Graphs (Ego-OMG) \cite{dessalene2020egocentric}, aggregating the representations from the Anticipation Module in producing representations for sequences of high-level states spanning large timespans of activity. Because of this we are able to abstract from contact derived representations to semantic modeling of the flow of activities. This also allows us to evaluate the utility of the Anticipation Module within the context of a full action understanding system.


The architecture of Ego-OMG consists of two streams. The first stream captures visual appearance and short term dynamics. This stream consists of a CSN \cite{Tran_2019_ICCV} a variant of the I3D Network \cite{carreira2017quo} making use of channel-wise group 3D convolutions. The second stream leverages the output of the Anticipation Module in modeling the temporal semantic structure of the activity being performed. The core of this second stream is a graph representation embedded into a vector space through use of a Graph Convolutional Network (GCN) \cite{kipf2016semi}.

Ego-OMG's graph representation is constructed as follows: transcripts of the activities from the training set are processed to produce a graph structure capturing the connections from state to state through actions. The nodes of this graph consists of state representations derivable from the Anticipation Module - categorical representations for the Active Object from the Contact Anticipation Network and the Next Active Object from the Next Active Object Network, modelling the left and right hands separately.

The CSN and GCN streams are then combined to produce an action prediction. 

We perform ablation studies over the Anticipation Module's representations, and through doing so determine which characteristics of those representations are responsible for their utility to action anticipation and prediction.


Using the representations produced by the full Anticipation Module we demonstrate state-of-the-art performance over the recent EPIC Kitchens Action Anticipation Challenge, achieving \nth{1} place on the EPIC Kitchens Action Anticipation Challenge unseen test set, and \nth{2} place on the seen test set, and outperform all previously published approaches without any use of ensembling, unlike many competing approaches.

The primary contributions of this work are:


\begin{itemize}

    
    \item A novel training signal for action understanding capturing information of time-to-contact between hands and objects, and segmentations of hands and objects. Over this signal we train the Anticipation Module, consisting of two networks which produce the following low level action representations:
    \begin{enumerate}
        \item Contact Anticipation Maps: pixel wise anticipated time-to-contact involving one of the left or right hands.
        \item Next Active Object Segmentations: segmentations localizing candidate Next Active Objects.
    \end{enumerate}


    


    
    
  \item  A surpassing of the state-of-the art with a full action understanding framework - Ego-OMG - built upon the proposed Anticipation Module, achieving \nth{1} and \nth{2} place on the unseen and seen test sets respectively of the EPIC Kitchens Action Anticipation Challenge.


\end{itemize}

The remainder of this paper is structured as follows: In Section \ref{sec:related_work} we provide an overview of related work; in Section \ref{sec:method} we detail our method; in Section \ref{sec:experiments} we describe our experiments and results; finally in Section \ref{sec:conclusion} we conclude.

\section{Related Work}
\label{sec:related_work}

\subsection{Action Anticipation and Prediction}
Action anticipation is the task of classifying future actions from observations that end before the actions begin. Action prediction is referred to in many works as "early action recognition": we adopt the nomenclature of \cite{kong2018human}, referring to the classifying of partially observed actions as action prediction. While the study of action recognition has received significant attention, the study of action anticipation and action prediction has only recently begun to attract more attention \cite{vondrick2015anticipating, gao2018ican, tanke2019human, abu2018will}, particularly in the egocentric setting \cite{furnari2019would, liu2019forecasting, nagarajan2020ego, ke2019time}.

\subsection{Egocentric Cues}

Previous works have demonstrated that exploiting hand motion and formation in various forms can improve action recognition performance \cite{bambach2015lending, urooj2018analysis}. Most previous action recognition frameworks incorporate hands by feeding hand detection patches \cite{bambach2015lending,yang2015grasp}, 3D joint pose estimations \cite{garcia2018first}, or both \cite{luvizon2017learning,tekin2019h+,baradel2018object}. Li et al. \cite{li2015delving} utilized the manipulation point, a 2D point in the image representing a point in reference to each of the hands, as an egocentric feature for action recognition. Fewer works attempt to utilize the hand trajectory as a cue. Liu et al. \cite{liu2019forecasting} propose motor attention, the anticipated future hand trajectory enacted throughout the performance of an action.

Rather than explicitly modelling future trajectories - which are inherently ambiguous - we focus on predicting the endpoint of the trajectories, terminating in contact with objects. For this, we leverage our Contact Anticipation Maps stacked through time. This history of Contact Anticipation Maps implicitly contains trajectory information.


\subsection{Active Objects}
\label{subsection:active_objects}

Anticipating future object interaction has been explored in many recent works. Furnari et al. \cite{furnari2017next} propose a method which relies on an object detector that exhaustively identifies a list of objects to track in a small sliding window - they feed each tracking trajectory to a random forest classifier to distinguish between 'active' and 'passive' trajectories. Nagarajan et al. \cite{nagarajan2019grounded} utilize pairs of inactive object images and videos of the corresponding objects in action, learning a mapping between the two to learn 'interaction hotspots', or regions of likely activity. Xiao et al. \cite{xiao2019reasoning} tackle the same task, proposing a novel architecture that utilizes objects to determine where actions are most likely to occur, and vice-versa.


In this work, we make a distinction with respect to these works as to the definition of an Active Object. Rather than refer to the object involved in the current action, we define an Active Object as the object presently in contact with a hand. This low-level definition of an Active Object better captures the objects involved in a given interaction. 


\begin{figure*}[t!]
\includegraphics[width=1\textwidth]{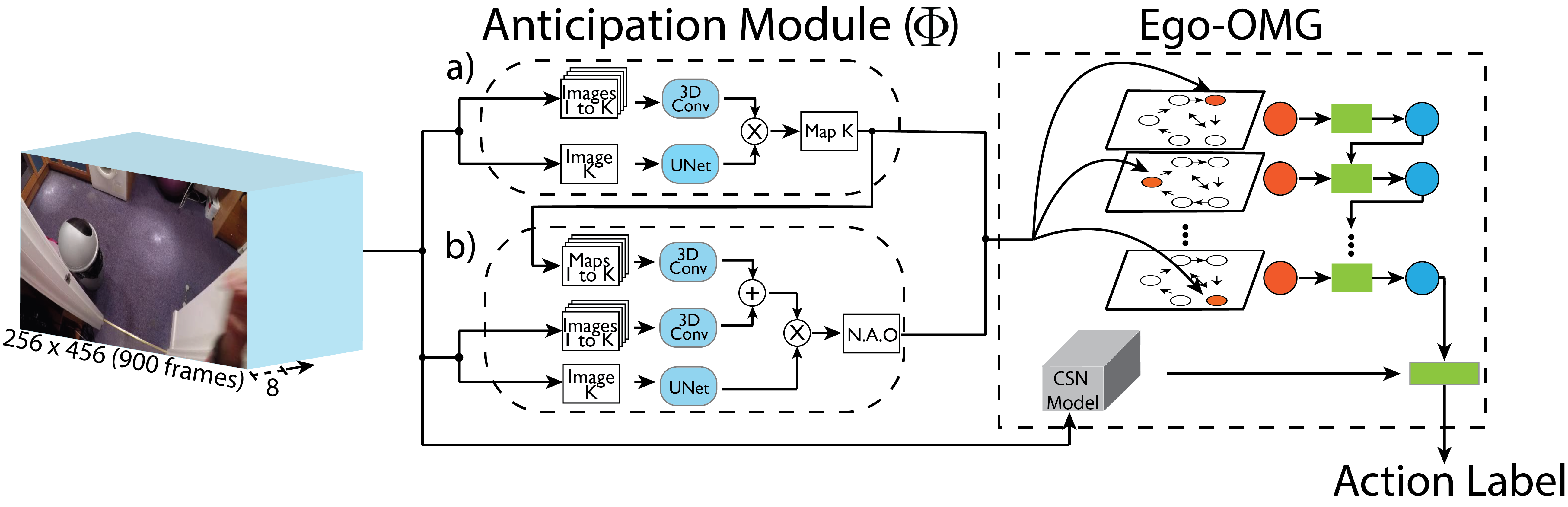}
  \caption{Overview of our proposed approach. The input video of $900$ frames is fed in sliding window fashion with windows of size $8$ to the Anticipation Module. Anticipation module $\Phi$ consists of two networks: a) The \textit{Contact Anticipation Network}, which outputs \textit{Contact Anticipation Maps (Map K)}, a representation which feeds into b) the \textit{Next Active Object Network}, producing a \textit{Next Active Object (N.A.O.) Segmentation}. The $\bigoplus$ denotes addition; $\bigotimes$ denotes multiplication. Refer to Sections \ref{subsec:contact_anticipation_network} and \ref{subsec:next_active_object_network} for the architectural details. The Anticipation Module $\Phi$'s output is in turn is fed into Ego-OMG, which in turn produces labels for action anticipation and prediction. }
  \label{fig:overall}
\end{figure*}


\begin{figure}[!t]
\centering
\includegraphics[width=0.45\textwidth]{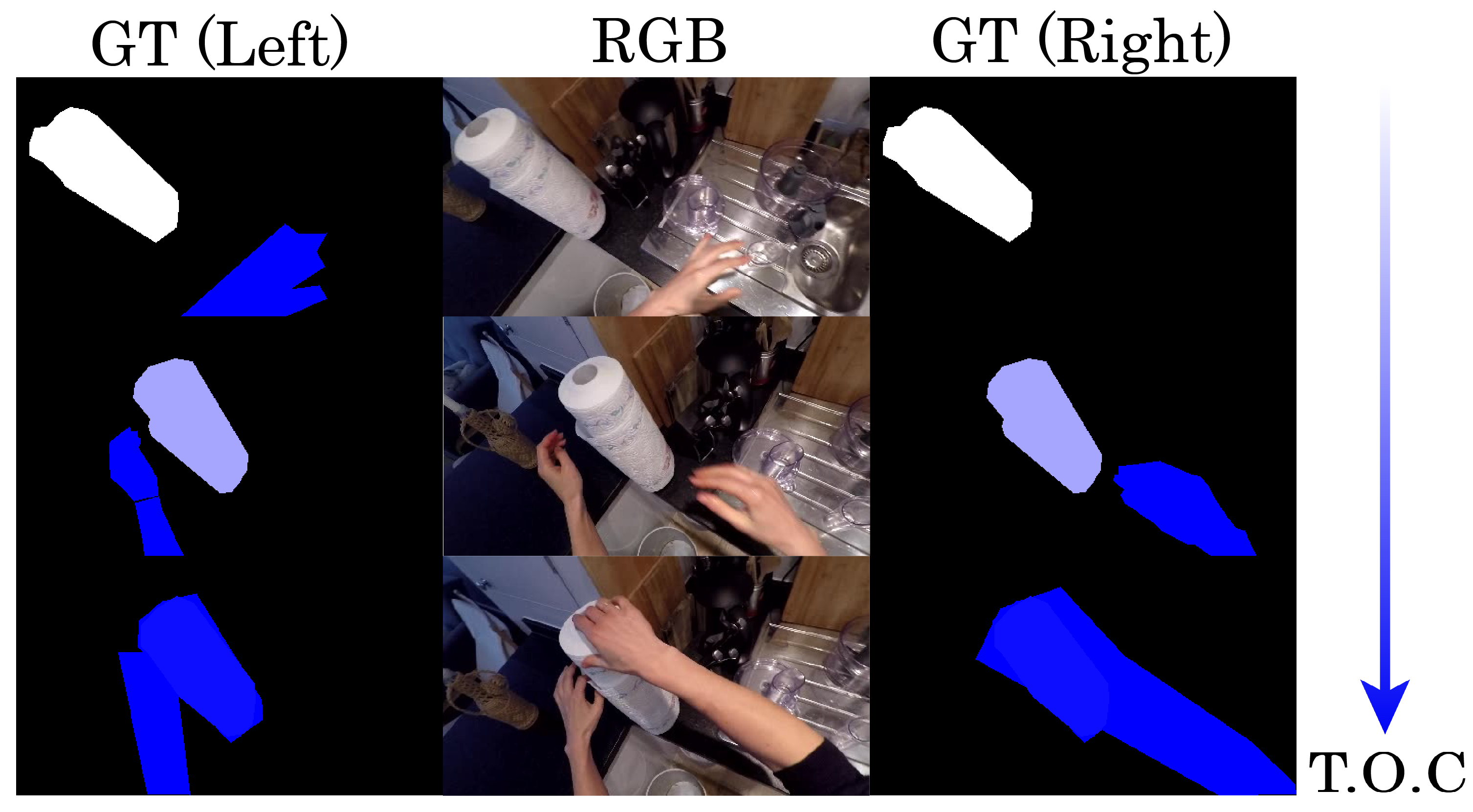}
\caption{Illustration of annotations added to a portion of the EPIC Kitchens dataset in construction of our augmented dataset. The left and right columns contain added annotations for the left and right hands, respectively. The middle column illustrates associated clip frames. The annotations consist of segmentations of hands and Active and Next Active Objects. Pixels belonging to (Next) Active Objects are assigned non-negative values relative to the time-of-contact (T.O.C). Colors vary from white to blue based on remaining time to contact, with values of 0 associated with both Next Active Objects and hands. Background pixels are colored black, represented with values of -1.}

\label{fig:dataset_figure}
\end{figure}

\subsection{Video Representation}
Typical works within action understanding involve two-stream architectures where the input to the network is RGB video fed to the network in parallel with pre-computed frames of optical flow \cite{carreira2017quo, simonyan2014very}. These approaches have achieved success in tasks where appearance and short-term motion is sufficient for the task at hand (i.e. action recognition) \cite{carreira2017quo}. However, it has been reported \cite{carreira2017quo, liu2019forecasting, li2018eye} that such methods do not transfer well to tasks such as action prediction or action anticipation. We find this understandable, as action anticipation requires reasoning about complex semantic cues that go beyond appearance.

Rather than simply represent the video as a stack of frames, it is desirable to capture the long-term semantics underlying the video observation of the activity. Recent works have proposed the enrichment of raw video features with graphs \cite{jain2016structural, Wang_2018_ECCV, nagarajan2020ego, soran2015generating}. Typically graph nodes represent detected objects, actors, or locations. Unlike other works that utilize an exhaustive list of entities, by restricting ourselves to the modelling of objects either currently or expected to be in contact with the hands, we are able to rule out 'background' objects that play no role in the actions involved, effectively using the hands as an attention mechanism. Furthermore, by aggregating contact based representations over larger timespans, we are able to model longer term structure of activity, whereas other approaches \cite{vondrick2015anticipating, gao2018ican, lan2015action} are centered on visual appearance and short term dynamics on the order of $1 - 2$ seconds.

\section{Method}
\label{sec:method}

\begin{figure*}[t!]
\includegraphics[width=1\textwidth]{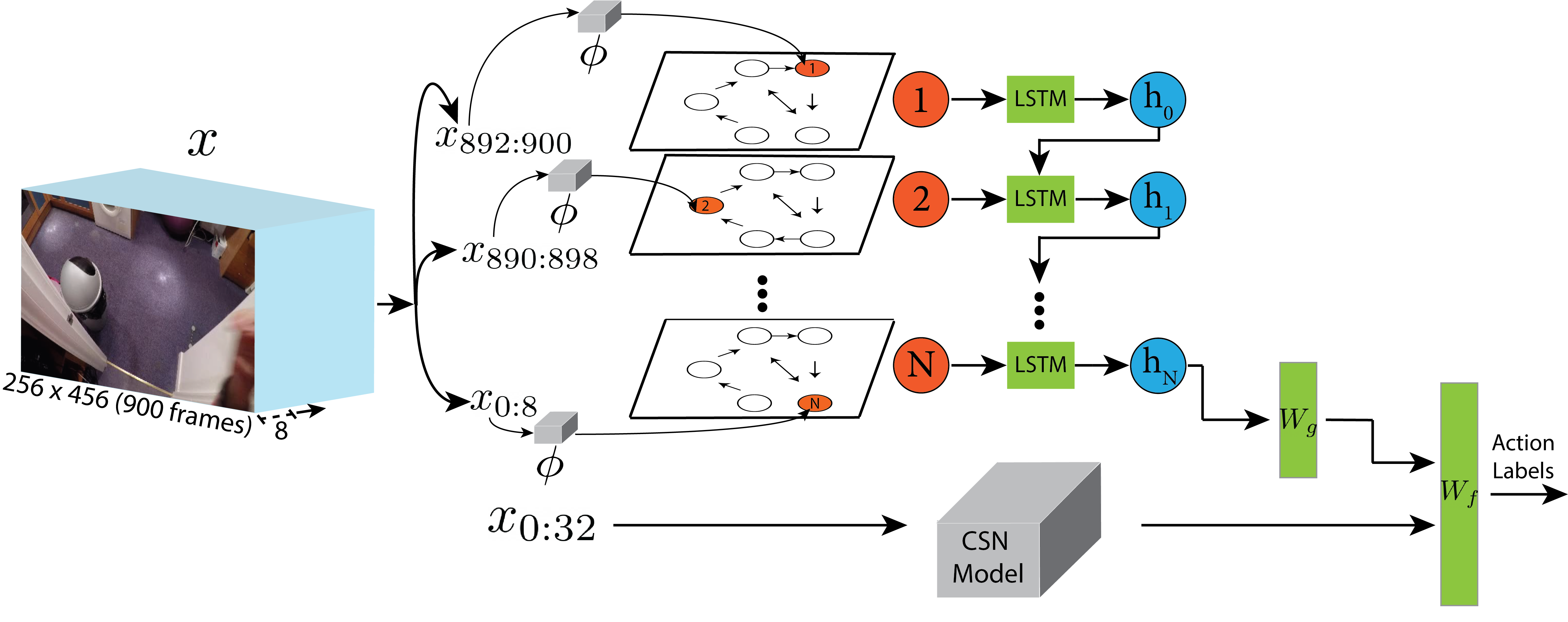}
  \centering
  \caption{Overview of Ego-OMG's architecture. Ego-OMG consists of two streams: 1) The top stream consists of the extraction of a discretized sequence of states from an unconstrained egocentric video clip $x$ of $900$ frames using the Contact Anticipation Network $\Phi$. The nodes predicted by $\Phi$ are embedded through GCN layers and then fed to an LSTM. This is then followed by a 1-layer MLP $W_g$ to generate softmax scores for the anticipated future action. 2) The second stream generates softmax scores for the anticipated future action through feeding a short history (the last 32 frames of $x$) of video to a CSN model. A 1-layer MLP $W_f$ processes the concatenated L2-normalized softmax scores to perform action anticipation and prediction.}
  \label{fig:architecture}

\end{figure*}

In this section we introduce our method for action understanding. Through contact and activity modeling our approach seeks to anticipate partially observed and/or near-future action. The structure of our approach is shown in Figure \ref{fig:overall}. Input video is fed first into the Anticipation Module - which we denote $\Phi$ - from whose output we produce symbolic state representations to be fed through Ego-OMG, which in turn anticipates partially observed and/or near-future action.


For the task of action anticipation, the observation of the video segment spans a range preceding the action start time $\tau_s$ by observation duration $t_o$, and ends $t_a$ seconds before $\tau_s$, where $t_a$ is the anticipation offset. In other words, input clips span from time $\tau_s - (t_o + t_a)$ seconds to end time $\tau_s - t_a$ seconds. For the task of action prediction, input clips span from time $\tau_s + p(\tau_f - \tau_s) - t_o$ to $\tau_s + p(\tau_f - \tau_s)$, where $\tau_f$ is the end time of the action and $p$ is the observable proportion of the clip containing the action to be predicted.



For our focus on hand-object contact in action modeling we devote the Anticipation Module. The Anticipation Module produces pixel-wise mappings of anticipated hand-object contact over the input. These mappings are divided into two types: Contact Anticipation Maps and Next Active Object Segmentations. The Contact Anticipation Network produces Contact Anticipation Maps, and is described in Section \ref{subsec:contact_anticipation_network}. The Next Active Object Network relies upon Contact Anticipation Maps for segmentation, producing Next Active Object Segmentations, and is described in Section \ref{subsec:next_active_object_network}. One advantage the Anticipation Module provides is that its mappings range over the near future action, and are not constrained to fixed anticipation time offsets as in several alternative action anticipation approaches \cite{vondrick2015anticipating,furnari2018leveraging}.

Training the Anticipation Module requires annotations for contact and localization of (next) active objects. To this end we augment the standard video data - in this work EPIC Kitchens - with temporal and segmentation information pertaining to contact. This process is described in Section \ref{subsec:dataset}.


We apply a Faster-RCNN \cite{ren2015faster} classifier over the maps produced by the Anticipation Module to produce symbolic states. These symbolic states capture characteristics of and relations between hands and objects in a compact representation. Symbolic state representations allow for easy use, and representation of state relations.

For our focus on temporal relational structure we devote Ego-OMG. Ego-OMG represents relations between action states across multiple time ranges, and uses these relations in contextualizing the present moment, and in projecting to near future action.

A natural formalism for representing temporal relations is a graph. We employ a graph in Ego-OMG to represent action state relations, and embed graph nodes into Euclidean space through use of word embeddings and a Graph Convolutional Network. Details of this process are described in Section \ref{sec:ego-omg}.

The remainder of Ego-OMG is as follows, and covered in detail in \ref{sec:ego-omg}. The sequence of states derived from the input is represented through the dynamics of an LSTM applied over embedded state representations. This LSTM allows projection into the near future. Finally, the anticipated action produced by this LSTM is joined by visual and short term dynamic information produced by a conventional 3D CNN. This component of Ego-OMG is swappable with alternative action understanding methods, making Ego-OMG complementary to many existing action understanding frameworks.

\subsection{Anticipation Module}
\label{sec:module}

In Section \ref{subsec:dataset} we describe the methods behind the collection of our dataset used for training the Anticipation Module, where the dataset consists of clips carefully selected from the EPIC Kitchens dataset. In Section \ref{subsec:contact_anticipation_network} we introduce the Contact Anticipation Network and in Section \ref{subsec:next_active_object_network} the Next Active Object Network, the two components that together form the Anticipation Module.
 
\subsubsection{Dataset}
\label{subsec:dataset}


We collect our dataset by organizing clips that correspond to point-to-point hand movements, where the hand involved and the Next Active Object are visible. The temporal boundaries of each clip are set such that clips begin when both the Next Active Object and the hand(s) targeting the object are visible, and end when the hand makes contact with the Next Active Object. As such, the lengths of the collected clips vary in the temporal dimension.


Rather than uniformly sample clips across all actions, we instead narrow our dataset to hand movement driven actions (i.e. \textit{take}, \textit{move}, \textit{cut}, \textit{open}) in the EPIC Kitchens dataset, as these actions each contain meaningful transitions in object status and encode the hand intentionality we wish to capture, making for a total of 2.1K randomly sampled clips with 252 unique object categories.


We croudsource our annotations on Amazon Mechanical Turk, asking workers to, for every 4 frames of a given clip, a) select the hand(s) involved in the given action and trace the Next Active Object, producing Next Active Object Segmentation masks $\{\Psi_l, \Psi_r\}$, and b) trace the left and right hands of the person and the objects held by each hand, creating contact segmentation masks $\{\Gamma_l, \Gamma_r\}$. We generate dense supervision of video using the forward and inverse warping of optical flow obtained from TVL1 \cite{perez2013tv}, projecting the annotations between annotated frames. That is, for flow displacements $u_{xy_1}, v_{xy_1} \in F_{t-1}^{t}$ and $u_{{xy}_2}, v_{{xy}_2} \in F_{t+1}^{t}$, values from each location $(x, y)$ in the segmentation masks are copied to pixel locations $(x + \frac{1}{2}(u_{xy_1} + u_{xy_2}), y + \frac{1}{2}(v_{xy_1} + u_{xy_2}))$, for warped subsequent frames, for flow frames $F$.

To generate the Contact Anticipation Map supervision training signal $C$, for an annotated frame taken at time $\tau$, $t_c$ seconds away from the time-of-contact, we retrospectively assign each pixel belonging to the annotated Next Active Object the value of $t_c$. Pixels corresponding to the body of the person or objects held in the hand at time $\tau$ are assigned self-contact values of $0$. All background pixels are populated with values of $-1$ and are not directly used during training. Figure \ref{fig:dataset_figure} provides an illustration of this process. 


To generate the Next Active Object binary segmentation masks $A$, we simply assign pixels belonging to the near-future contacted object values of $1$, and assign values of $0$ to all other pixels.

The Contact Anticipation Maps and the Next Active Object Segmentations each consist of two separate pixel-level channels $C = \{C_r, C_l\}$ and $\Psi = \{\Psi_r, \Psi_l\}$ respectively, for the right and left hands. In clips involving bi-manual manipulation, the Contact Anticipation Maps for the channels of each hand differ due to the different timings underlying the movement of each hand. However, the Next Active Object masks are shared between the channels of each hand, or $\Psi_l = \Psi_r$. 

\subsubsection{Contact Anticipation Network}
\label{subsec:contact_anticipation_network}

The Contact Anticipation Map predictions require the modelling of short-term dynamics for capturing the underlying hand trajectory and the localization of boundaries pertaining to hands and objects in contact. To capture both, we devise a custom two-stream architecture: One stream consisting of 3D Convolutions applied over the input video for modelling short-term dynamics, and another consisting of a U-Net stream applied over a single frame belonging to the end of the observation for capturing more precise hand segmentations. See $\Phi$ in Figure \ref{fig:overall} for an illustration.

The Contact Anticipation Network (see $\Phi$ part a in Figure \ref{fig:overall}) takes a stack of $8$ sequential RGB frames and outputs four channels: Two pixel-level regression outputs $\{D_l, D_r\}$ corresponding to the estimated remaining time-to-contact for each pixel in the image, and two soft segmentation maps. We threshold the soft segmentation maps to arrive at binary segmentation masks $\{\hat{\Gamma}_l, \hat{\Gamma}_r\}$, containing pixel-level segmentation masks of hands and objects in contact with the hand. The two output channels in both cases are for the predictions separately designated for the left and right hand, respectively, each of size $(128, 228)$.

The 3D Convolutional stream is a standard 3D ResNet50 architecture, where the backbone network from \cite{carreira2017quo} is utilized. It consists of 5 successive 3D Convolutional layers, where the first and third layers are followed by 3D Max Pooling operations. The UNet stream is composed of the exact architecture proposed in \cite{ronneberger2015u}, where a contractive path (two 2D Convolutions followed by a 2D Max Pooling operation) is followed by the expansive path (2D Transposed Convolution layers followed by 2D Convolutions). The network is trained using ADAM with a learning rate of $0.0001$ and a decay of $5\mathrm{e}{-6}$. We apply ResNet-style normalization, and augment the input RGB video with standard crops, flips, and color jitters.


There are two loss components used in training the Contact Anticipation Network. The first component, $L_{MAE}$, is the pixel-wise mean average error between predictions $\{D_l, D_r\}$ and ground truth $\{C_l, C_r\}$, only over pixel locations $(x, y)$ where $C_{l_{xy}} > 0$ and $C_{r_{xy}} > 0$. In other words, this loss component is only computed over pixels belonging to the Next Active Object; other pixels do not have time-to-contact annotations, and so they are ignored. The second component, $L_{BCE}$, is the binary cross entropy loss between the predicted soft segmentation maps and contact segmentation masks $\{\Gamma_l, \Gamma_r\}$. The loss used to train the system is as follows: $L = L_{BCE} + \gamma L_{MAE}$, where $\gamma = 0.2$.

We predict pixels of contact $\{\hat{\Gamma}_l, \hat{\Gamma}_r\}$, where $\hat{\Gamma}_{s_{xy}} = 1$ for hand side $s \in (l, r)$ if pixel location $(x, y)$ corresponds to a hand or object in contact and $\hat{\Gamma}_{s_{xy}} = 0$ otherwise. To arrive at the Contact Anticipation Maps, we superimpose the predicted pixels of contact $\{\hat{\Gamma}_l, \hat{\Gamma}_r\}$ over the regressed time maps $\{D_l, D_r\}$, for each hand side, to arrive at Contact Anticipation Maps $\hat{C}$, as follows:


\[
  \hat{C}_{l_{xy}} =
  \begin{cases}
       0 & \text{if $\hat{\Gamma}_{l_{xy}} = 1$} \\
       D_{l_{xy}} & \text{if $\hat{\Gamma}_{l_{xy}} = 0$} \\
  \end{cases}
\]

\[
  \hat{C}_{r_{xy}} =
  \begin{cases}
       0 & \text{if $\hat{\Gamma}_{r_{xy}} = 1$} \\
       D_{r_{xy}} & \text{if $\hat{\Gamma}_{r_{xy}} = 0$} \\
  \end{cases}
\]

The final Contact Anticipation Maps $\{\hat{C_l}, \hat{C_r}\}$ are fine-grained distributions of non-negative continuous values for each pixel that represents the estimated time of contact. Each of the channels associated with the left and right hand are of size $(128, 228)$.







\subsubsection{Next Active Object Network}
\label{subsec:next_active_object_network}

As illustrated in $\Phi$ part b of Figure \ref{fig:overall}, the $8$ frame RGB video and $8$ frame Contact Anticipation Map history are fed in parallel through 3D Convolutions, after which a summation over the stream is performed. Additionally, the final frame of the $8$-frame input is fed into a U-Net architecture in order to capture more precise object segmentations. Next, a pixel-wise multiplication between the resultant feature map from the 3D Convolutional streams and the output of the U-Net model is performed. The result of this multiplication is fed through sigmoid activation units, producing soft segmentation maps for the right and left hands, which are binarized using a threshold of $0.15$ to arrive at $\{\hat{\Psi}_r, \hat{\Psi}_l\}$.

Each of the two 3D Convolutional streams have architectures identical to those used in the 3D Convolutional stream in \ref{subsec:contact_anticipation_network}. Likewise, the U-Net stream is identical to that of \ref{subsec:contact_anticipation_network}. The final output of the combined streams is of size $(128, 228)$. The network is trained using ADAM with a learning rate of $0.0001$ and a decay of $5\mathrm{e}{-6}$. We utilize a weighted binary cross entropy loss function between ground truth $\{\Psi_l, \Psi_r\}$, and predictions $\{\hat{\Psi}_l, \hat{\Psi}_r\}$ with a weight value of $2.0$ chosen to overcome the foreground/background class imbalance in the ground truth Next Active Object masks of the collected dataset.

To avoid overfitting on the Contact Anticipation Map stream, multiplicative Gaussian Noise sampled independently over each pixel is applied over the output of the Contact Anticipation Map stream, adding $\hat{C_i} \odot Z_i$ where $Z_i = \mathcal{N}(\mu,\,\sigma^{2})$, where $\odot$ is the element-wise Hadamard product. This augmentation captures the inherent ambiguity of anticipating contact; there is little ambiguity in predicting the time values of pixels belonging to hands or contacted objects due to their proximity (by definition having time-to-contact of $0$), while there is increasing ambiguity in predicting time-to-contact for objects the further from the hands they are. We apply ResNet-style normalization, and augment the input RGB video with standard crops, flips, and color jitters.





\subsection{Ego-OMG}
 \label{sec:ego-omg}

As illustrated in Figure \ref{fig:architecture}, we feed input video $x$ into the Anticipation Module $\Phi$, whose purpose is to predict and anticipate hand object contacts. Current predicted and future anticipated contact is represented through a $4$ channel output, consisting of two contact segmentation masks $\{\hat{\Gamma}_{t_r}, \hat{\Gamma}_{t_l}\}$ produced by the Contact Anticipation network and two object segmentation masks $\{\hat{\Psi}_{t_r}, \hat{\Psi}_{t_l}\}$ produced by the Next Active Object network, where $\hat{\Psi}_{t_r}$ and $\hat{\Psi}_{t_l}$ denote the predictions of the Next Active Object, and $\hat{\Gamma}_{t_r}$ and $\hat{\Gamma}_{t_l}$ denote the objects detected to be presently in contact with the hand, both for the right and left hands respectively. We classify each segmentation frame with a pre-trained Faster-RCNN \cite{ren2015faster} model, arriving at predicted object classes $o_t = \{\psi_{t_r}, \psi_{t_l}, \gamma_{t_r}, \gamma_{t_l}\}$. We note that for the purposes of this work we predict up to $1$ object each for $\psi_{t_r}$, $\psi_{t_l}$, $\gamma_{t_r}$, and $\gamma_{t_l}$. This limitation prevents us from modelling scenarios where multiple objects are held by the same hand for tasks requiring dexterous manipulation.

In practice, while the Contact Anticipation Network succeeds at localizing contacted objects, the classifier tends to mis-classify currently held objects due to the severe occlusion imposed by the hand, especially for small objects like scissors and utensils. Therefore, in building the graph we impose the constraint that every object currently contacted by each hand \textit{must have been anticipated} at some previous instance in time, before the presence of occlusion. In classifying the objects currently in contact with the hand, we take the intersection of top-5 object class predictions for that object with the object classes previously predicted in anticipation over the past $100$ frames ($7$ seconds).

In this section we define Ego-OMG, a two-stream architecture dependent on a novel graph representation $G$ that consists of a structured sequence of high-level states extracted from videos belonging to the EPIC Kitchens dataset. The graph $G$ contains two types of nodes: 1) nodes spanning current contact and forecasted contact of hands and objects, which are produced by the anticipation module in (ref), and 2) nodes corresponding to action labels in the EPIC Kitchens dataset. The graph $G$ consists of edges connecting state-to-state transitions and state-to-action co-occurence.

Section \ref{subsec:joint} details the two-streams of Ego-OMG: the first modeling temporal relations and context, and the second modelling visual appearance and short-term dynamics through use of a 3D CNN. Section \ref{subsec:construction} explains the construction of the graph of Ego-OMG used in producing structured video representations. 


\subsubsection{Joint Architecture}
\label{subsec:joint}

The architecture of Ego-OMG is shown in Figure \ref{fig:architecture}. Input consists of a single clip spanning 30 seconds - or 900 frames. The output consists of a logit layer predicting the class of the action $\tau_a$ seconds after the end of the observation. The architecture is comprised of two streams: One modeling the appearance and short term dynamics of the last few seconds of the clip; the other modeling hand dynamics and long-term semantic temporal relations.

In the first stream, we model appearance and short-term dynamics with a Channel-Separated Convolutional Network (CSN), a 3D CNN factorizing 3D convolutions in channel and space-time in similar fashion to Xception-Net \cite{chollet2017xception} which factorizes 2D convolutions in channel and space. The weights are pre-trained on the largescale IG-65M video dataset \cite{ghadiyaram2019large}. The network takes as input $32$ frames of size $256\times256$. We apply horizontal flipping, color jittering and random crops during training, with centered crops during testing. The model is trained using SGD with a batch size of $16$, a learning rate of $2.5 \times 10^{-3}$ and a momentum of $0.9$.

In the second stream we model dynamics of interactions between hands and objects, as well as longer term temporal semantic relations between the actions of the activity. We capture this structure in the form of a graph, described in detail in Section \ref{subsec:construction}. After computing the graph, we feed it through two graph convolution layers of hidden layer size $256$ and $128$ respectively. Note our application of the GCN is transductive; it is applied on a single, fixed graph consisting of all nodes seen during train \textit{and} test time beforehand. We feed the sequence of node embeddings obtained by the GCN into an LSTM \cite{hochreiter1997long}. At test time, we convert an input video of $900$ frames to a sequence of states and from each state's respective node embedding $g_n$ for $n \in N$, we aggregate the state history with a 1-layer LSTM. From the LSTM's final hidden state $h_N$, we apply a 1-layer MLP $W_g$ to classify the next most likely action. The LSTM carries hidden states of size $128$. A batch size of 16 and a learning rate of $7 \times 10^{-5}$ is used with ADAM optimizer and a cross entropy loss function. Training achieves fast convergence, reaching peak top-1 action anticipation and action prediction accuracy after $5$ epochs or roughly $0.25$ hours of training on a NVIDIA GeForce GTX 1080 GPU.

We concatenate the L2-normalized softmax scores from each respective stream, freezing the two sub-networks while training a 1-layer MLP $W_f$ with a batch size of $16$ and learning rate of $0.01$ on top of the joint softmax scores to classify the next most likely action. We find a late fusion approach provides slight benefits in practice as opposed to an early fusion of the two streams, likely due to the different learning dynamics of the individual streams. Inference times are dominated by the CSN model.
\label{subsec:action_understanding}

\subsubsection{Graph Construction}
\label{subsec:construction}

We have a set of $K$ training videos. To detect the objects involved in interaction, which are needed to build the graph, we utilize both sub-components of the Anticipation Model $\Phi$, described in subsections \ref{subsec:contact_anticipation_network} and \ref{subsec:next_active_object_network}. The Anticipation Module $\Phi$ iterates over each video using a sliding window with an $16$-frame width, sampling every $2$ consecutive frames with a stride of 2. Feeding each of $4$ output channels of $\Phi$ to the object classifier then produces detections $O_i = \{o_1, o_2, ..., o_{T_i/2}\}$ for video $i$, where $T_i$ is the frame count of video $i$. From the per-frame predictions of the object classes $o_i$, we suppress consecutive duplicate predictions arriving at non-consecutively repeating states $S_k = \{s_1, s_2, ..., s_n\}$, a sequence where temporal order is preserved.


With the input to graph construction defined, we now consider the graph $G = (V, E)$, where $E$ consists of the set of all edges, and $V$ consists of the set of all nodes. $V = \{V_s, V_a\}$ consists of nodes of two types: state nodes, and action nodes. State nodes consist of the union of all $S_k$, that is: $V_s=\bigcup_{k=1}^{K} S_k$, and action nodes $V_a$ consist of the set of all action classes $a_i \in A$, where $A$ is the set of all actions. In doing so, we represent both states and actions in graph $G$.

We construct the adjacency matrix as follows. Each node has an edge connecting it to itself: $e_{ii} \in E$ for $1 \leq i \leq |V|$ with weight $1$. We add weighted directed edges $e_{ij} \in E$ for consecutive states $s_i$ and $s_j$ for $0 \leq i < n$ and $j = i + 1$, where the weight $\sigma_{ij}$ is transition probability $p(s_{i+1} | s_i)$ where transition probabilities are observed from transitions in state sequences $S_k$ for all $k \in K$. We also add weighted directed edges between states and actions by adding weighted edge $e_{ij} \in E$ if action $i$ takes place within the timespan of state $s_j$, where weight $\sigma_{ij}$ is equal to $p(a_i | s_j)$.

Graph $G$ has a total number of nodes equal to the number of unique states $z = \abs{S} + \abs{A}$, where $S$ is the set of unique states and $A$ is the set of annotated actions. Let $X \in R^{z \times m}$ be a matrix containing all $z$ nodes with their corresponding features of dimension $m$. Rather than set $X$ to identity matrix $I$, we initialize each node with feature embeddings extracted from a pre-trained GloVe-600 model \cite{pennington-etal-2014-glove}. When representing states $s \in S$, we average the feature embeddings from each object noun in $s$. When representing actions $a \in A$, we average the embeddings for the verb and noun embeddings. We find that utilizing pretrained word embeddings for $G$ results in substantial performance gains over using $X = I$.

We feed the weighted adjacency matrix and $X$ as input into the GCN as described in Section \ref{subsec:joint}.

\section{Experiments}
\label{sec:experiments}

Throughout these experiments we evaluate the performance of the proposed models for action anticipation, action prediction, and Next Active Object prediction. We also perform ablations over the components of the Anticipation Module to understand their respective contributions to the success of the entire framework.



\begin{table}[b]
   \centering
        \begin{tabular}{ r|c|c|c }
        \multicolumn{1}{r}{}
         &  \multicolumn{1}{c|}{Method}
         & \multicolumn{1}{c}{Top-1}
         & \multicolumn{1}{c}{Top-5}\\
        \hline{}
         & 2SCNN (RGB) \cite{shou2016temporal} & 4.32 & 15.21\\
         & TSN (RGB) \cite{wang2016temporal} & 6.00 & 18.21\\
         & TSN + MCE \cite{furnari2018leveraging} & 10.76 & 25.27 \\
        S1 & RULSTM\cite{furnari2019would} & 15.35 & 35.13 \\
         & Camp. et al. \cite{camporese2020knowledge} & 15.67 & \textbf{36.31} \\
         & Liu et al. \cite{liu2019forecasting} & 15.42 & 34.29 \\
         & \textbf{Ours} & \textbf{16.02} & 34.53 \\
        \hline{}
         & 2SCNN (RGB) \cite{shou2016temporal} & 2.39 & 9.35 \\
         & TSN (RGB) \cite{wang2016temporal} & 2.39 & 9.63 \\
         & TSN + MCE \cite{furnari2018leveraging} & 5.57 & 15.57 \\
        S2 & RULSTM \cite{furnari2019would} & 9.12 & 21.88 \\
        & Camp. et al. \cite{camporese2020knowledge} & 9.32 & 23.28 \\
         & Liu et al. \cite{liu2019forecasting} & 9.94 & 23.69 \\
         & \textbf{Ours} & \textbf{11.80} & \textbf{23.76} \\
        \end{tabular}
       \caption{Action anticipation results on the EPIC Kitchens test set for \textit{seen} kitchens (\textbf{S1}) and \textit{unseen} kitchens (\textbf{S2}) during the EPIC Kitchens Action Anticipation Challenge. Only published submissions are shown.}
      \label{table:comparison_other_methods}
\end{table}

\begin{table*}
\centering
\caption{Action anticipation and action prediction accuracy results over validation set for CSN stream, GCN stream and CSN + GCN stream over varying anticipation times $\tau_a$ seconds and varying observation rates $p$.}
\label{table:anticipation_time_vary}
\begin{tabular}{ |l*{12}{c}r| }
\hline
& \multicolumn{6}{ c| }{Action Anticipation ($\tau_a$)} & \multicolumn{6}{ c| }{Action Prediction ($p$)}\\
\hline
\multicolumn{1}{|r}{} & 5 & 2.5 & 1.5 & 1 & 0.5 & \multicolumn{1}{r}{0} & 
\multicolumn{1}{|r}{} & 12.5 & 25 & 50 & 75 & \multicolumn{1}{r|}{90} \\
\hline
CSN &  6.49 & 11.39 & 14.09 & 15.50 & 18.61 & 19.37 & \multicolumn{1}{|r}{} & 24.23 & 26.49 & 30.72 & 31.08 & \multicolumn{1}{r|}{31.30} \\ \hline
GCN &  9.05 & 10.47 & 11.31 & 12.81 & 13.76 & 14.56 & \multicolumn{1}{|r}{}  & 14.83 & 15.44 & 15.70 & 15.88 & \multicolumn{1}{r|}{16.01} \\ \hline
CSN + GCN & \textbf{9.44} & \textbf{15.01} & \textbf{17.02} & \textbf{19.20} & \textbf{20.29} &  \textbf{21.89} & \multicolumn{1}{|r}{}   &  \textbf{26.01} & \textbf{28.33} &  \textbf{31.14} & 31.19 & \multicolumn{1}{r|}{31.42}\\ \hline
RULSTM \cite{furnari2019would} &  6.98 & 10.92 & 12.31 & 12.69 & 16.98 & 18.21 & \multicolumn{1}{|r}{}  & 24.48 & 27.63 & 30.93 & \textbf{33.09} & \multicolumn{1}{r|}{\textbf{34.07}} \\ \hline
\end{tabular}
\end{table*}

\subsection{EPIC Kitchens Action Anticipation Challenge}
The protocol behind the EPIC Kitchens Action Anticipation Challenge is to set the anticipation time $\tau_a$ to $1$ second. While there are 44 participants in the challenge, we report our results alongside the top $3$ published submissions (\textbf{RULSTM} \cite{furnari2019would}, \textbf{Camp et al.} \cite{camporese2020knowledge}, \textbf{Liu et al.} \cite{liu2019forecasting}) and include the benchmarked action anticipation results from the EPIC Kitchens dataset release (\textbf{2SCNN} \cite{shou2016temporal}, \textbf{TSN (RGB)} \cite{wang2016temporal}, and \textbf{TSN + MCE} \cite{furnari2018leveraging}). See the Supplementary Material for details of each baseline.

Table \ref{table:comparison_other_methods} shows our results over the test set (S1) where scenes appear in the training set and over the test set (S2) where scenes are \textbf{not} included in the training set. We are \nth{2} place in S1, beating previous state-of-the-art methods by a margin of $.35\%$ and \nth{1} place in S2, beating previous state-of-the-art methods by a margin of $1.86\%$. We posit that the reason for Ego-OMG's notable outperformance w.r.t previous methods in S2 is because previous methods rely heavily on visual appearance and are more likely to fail when testing on unseen kitchens which likely include objects of previously unencountered appearance; Ego-OMG's GCN stream on the other hand only models objects of interaction, ignoring the diverse, cluttered, backgrounds that typically make up everyday kitchen environments.

The final column of Table \ref{table:comparison_other_methods} contains the evaluation of all methods with Top-1 evaluation and Top-5 evaluation - a prediction is correct w.r.t. Top-5 evaluation if the ground truth is included in the Top-5 predicted actions. In our approach we utilize the same model for Top-1 and Top-5 evaluation measures; other approaches may train separate models for the two evaluation measures. Furthermore, we note that the other methods incorporate distinct training mechanisms for top-5 action anticipation \cite{furnari2018leveraging, camporese2020knowledge}.


We stress that the CSN stream can be swapped with any of the architectures listed above; as the CSN stream is outperformed by \textbf{RULSTM}, \textbf{Camp. et al.}, and \textbf{Liu et al.}, better performance could be expected from Ego-OMG with the incorporation of any one of these architectures. In addition, we do not perform any form of ensembling in our submission.

\subsection{Action Anticipation and Prediction}
We evaluate our approach over the EPIC Kitchens dataset on the tasks of action anticipation and action prediction over varying anticipation times $\tau_a$ (action anticipation) and varying observation ratios $p$ (action prediction). The tasks are detailed in Section \ref{sec:method}. See Table \ref{table:anticipation_time_vary} for results. The purpose of these experiments is to analyze the performance of our approach and its components over degrading anticipation times and varying action observation ratios.

We vary the anticipation time $\tau_a$ from $0$ seconds (predicting the action class immediately before its start) to $5$ seconds (predicting the action class $5$ seconds before its start). We perform action prediction at the following observation ratios $p$: $12.5\%$, $25\%$, $50\%$, $75\%$ and $90\%$. 

We compare our approach to the state-of-the-art RULSTM \cite{furnari2019would} work due to its state-of-the-art performance over published methods in both action anticipation and action prediction, making RULSTM the optimal baseline.


For action anticipation, we note the performance of our approach degrades gracefully as anticipation time $\tau_a$ increases. As the anticipation time increases, the performance of the CSN stream drops off considerably, to the point where at $\tau_a = 5$ seconds, the GCN stream outperforms the CSN stream by a large margin of $2.56\%$. Full Ego-OMG outperforms each of its streams individually over all $\tau_a$.

For action prediction, we observe diminishing gains of the GCN stream's contribution to the performance of Ego-OMG as the observation ratio $p$ increases, to the point where RULSTM outperforms our approach after $p = 50\%$. We also observe the performance of our approach at observation ratios $p = 50\%$ and $p = 90\%$ are very close. These findings lead us to the conclusion that the strength of our approach lies in its anticipatory capabilities, and that our approach is not particularly better suited for the action recognition setting over other methods focusing on visual appearance and short-term dynamics. However, we hypothesize that the incorporation of flow into our approach would be of great benefit for the action recognition setting.

\subsection{Next Active Object}
\begin{figure}[b!]
\includegraphics[width=0.49\textwidth]{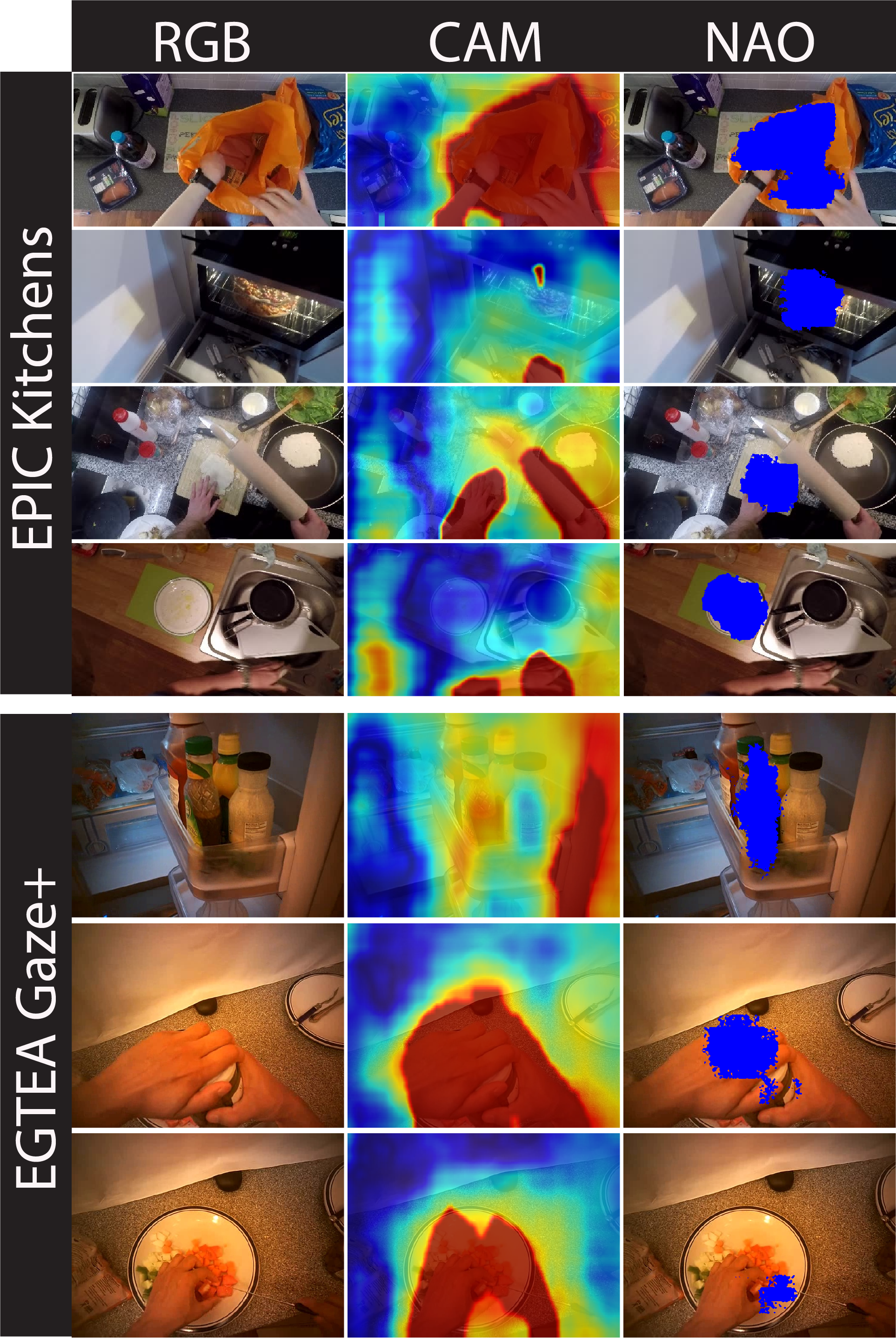}
  \centering
  \caption{The Anticipation Module outputs Contact Anticipation Maps (second column) and Next Active Object segmentations (third column). The Contact Anticipation Maps contain continuous values of estimated time-to-contact between hands and the rest of the scene (visualizations varying between red for short anticipated time-to-contact, and blue for long anticipated time-to-contact). The predicted Next Active Object segmentations contain the object of anticipated near-future contact, shown in blue in the third column. Predictions are shown over the EPIC Kitchens and the EGTEA Gaze+ datasets.}
  \label{fig:qualitative}

\end{figure}

In this section we evaluate the performance of the Anticipation Module on the prediction of the Next Active Object. We conduct two evaluations: The first being the evaluation of the localizations produced by the Anticipation Module, and the second being the evaluation of the classification over those produced localizations. Both evaluations are performed frame-wise over the test set of the augmented dataset. To illustrate the generalization of the Anticipation Module to other egocentric activity datasets, we provide the outputs of the Anticipation Module over the EGTEA Gaze+ dataset on Google Drive\footnote{\vspace{-1cm} Google Drive link at : https://drive.google.com/drive/folders/1AIZ93d37g0mJaHclANhXYVyp2jFQtfCS?usp=sharing}, where the Anticipation Module was trained over EPIC Kitchens. In addition, outputs of the Anticipation Module over both EPIC Kitchens and EGTEA Gaze+ are shown in Figure \ref{fig:qualitative}.



\begin{table}[b]
\centering
\begin{tabular}{l*{4}c}
\toprule
Evaluation & Jaccard \\ 
\midrule
Obj-Tracker       & 0.028    \\
DeepGaze II        & 0.051     \\ 
I3D-GradCam       & 0.079       \\ 
Center Bias   & 0.088  \\
Ours w/o CAM       & 0.169      \\
Ours   & \textbf{0.194}        \\ 
\bottomrule
\end{tabular}
\vspace{1em}
\caption{Evaluation of localizations produced by the Next Active Object predictions. Contact Anticipation Maps are referred to as CAM in the table.}
\label{table:next_active_object_prediction_results}
\end{table}

\subsubsection{Localization}
\label{subsection:localization}

The first set of Next Active Object evaluations is performed with respect to the ground truth segmentation masks included in the augmented dataset described in \ref{subsec:dataset}. We report Jaccard similarity as our evaluation measure. We provide an evaluation comparing baselines and an ablated and non-ablated implementation of our approach:




\begin{itemize} 
	\item \textbf{Center Bias} relies on the assumption that the Next Active Object most commonly appears near the center of the frame, and instantiates a fixed Gaussian at the center of each image of size $(55, 55)$ to represent the Next Active Object.
	\item \textbf{I3D-GradCam} trains an I3D model to perform action anticipation ($\tau_a$ sampled uniformly between $0$ to $2$ seconds) over the entire EPIC-Kitchens dataset, applying standard Grad-Cam \cite{selvaraju2017grad} over the trained network to generate heatmaps containing the Next Active Object. We mask out the hands of the actor from the heatmaps using ground truth hand segmentations from \ref{subsec:dataset} to better localize the Next Active Object.
	\item \textbf{DeepGaze II} \cite{Kummerer_2017_ICCV} is a pre-trained state-of-the-art model performing saliency prediction over each image in the collected dataset.
	\item \textbf{Obj-Tracker} is a re-implementation of \cite{furnari2017next}. The SORT \cite{bewley2016simple} tracking algorithm is applied over the detections obtained by Faster-RCNN pretrained over EPIC-kitchen dataset. Objects tracked for less than $20$ frames are dropped, and remaining trajectories are classified as either 'active' or 'inactive'. As predictions are bounding boxes, we convert the ground truth segmentation masks to bounding box format.

	\item \textbf{Ours w. and w/o CAM} are our proposed approaches with and without the incorporation of Contact Anticipation Maps in the Next Active Object Network to demonstrate the utility of Contact Anticipation Maps. Other approaches do not model left and right hands separately, and so for fair comparison we collapse our model's binarized two channel output for the left and right hands, into one channel.
\end{itemize}

\begin{table}
\centering
\begin{tabular}{l*{6}c}

\toprule
Evaluation & Top-1 & Top-5 & \\ 
\midrule
Obj-Tracker       & 1.00       & 5.70           \\
I3D Classifier   & 11.90      & 31.94         \\
RULSTM & 15.07  & \textbf{39.88} & \\
Ours        & \textbf{18.26}      & 39.67      \\ 
\bottomrule
\end{tabular}
\vspace{1em}
\caption{Evaluation of classification accuracy with respect to the Next Active Object predictions.}
\label{table:next_active_object_class_results}
\vspace{-.7cm}
\end{table}

Our approach including the Contact Anticipation Maps outperforms all baselines and the ablation by large margins. We attribute this to its rich encoding of hand trajectory.  The performance of \textbf{Obj-Tracker} on the EPIC Kitchens dataset is poor compared to its performance in \cite{furnari2017next} over the AVL dataset. We observe that the tracker fails consistently in tracking objects over timespans exceeding $~.5$ seconds. For the frames where the objects are tracked, next active object localization is reported.

\subsubsection{Classification} The second set of Next Active Object evaluations is performed with respect to the annotated object classes available from the EPIC Kitchens dataset. We provide an evaluation comparing three baselines and our approach:

\begin{itemize}
	\item \textbf{I3D Classifier} trains an I3D model over the action segments of the EPIC Kitchens dataset, using the object noun labels as ground truth to perform end-to-end classification of the future object of interaction with action anticipation offset $\tau_a$ sampled uniformly between $0$ to $2$ seconds.
	\item \textbf{Obj-Tracker} is a re-implementation of \cite{furnari2017next}. See the baseline description at Section \ref{subsection:localization} for more.
	\item \textbf{RULSTM} \cite{furnari2019would} is a state-of-the-art architecture with two separate LSTMs and an attention formulation applied over features obtained from RGB, flow, and object detections. This, like \textbf{I3D Classifier}, is trained over the action segments from the EPIC Kitchens dataset.
	\item \textbf{Ours} computes the Intersection-over-Union values between the bounding boxes produced by a Faster-RCNN model trained over EPIC Kitchens and the predicted Next Active Object segmentation. The object category associated with the highest IoU is used as the prediction.
\end{itemize}

Our approach outperforms the \textbf{I3D Classifier} and \textbf{RULSTM} baselines at Top-1 Next Active Object prediction by large margins, and is only marginally outperformed by \textbf{RULSTM} at Top-5 Next Active Object prediction. We note that \textbf{RULSTM} and \textbf{I3D Classifier} are trained on input clips with action segments of temporal boundaries defined by the EPIC Kitchens annotations, whereas \textbf{Obj-Tracker} and \textbf{Ours} were trained on input clips with temporal boundaries defined by the augmented dataset. Also, \textbf{RULSTM} and \textbf{I3D Classifier} are trained over 10X the number of clips that our Anticipation Module is trained over (28K vs 2.1K respectively). We expect our approach could benefit from a larger augmented dataset.




\subsection{Anticipation Module Ablations}
We perform several ablation studies over the components of the Anticipation Module to understand their contributions towards their performance of the GCN stream in Ego-OMG for the tasks of action anticipation ($\tau_a = 1$ second) and prediction ($p = 0.25$).

\subsubsection{Hand Representation}
We compare the effects of having the Anticipation Module produce localizations for each hand individually vs. jointly. Our approach consisting of Active and Next Active Object predictions for each of the two hands is compared with \textbf{Joint}, where we treat both hands as one entity by superimposing the binarized segmentation channels produced for each hand. This ablation evaluates the extent to which distinctly modelling left and right hands benefits action anticipation and prediction. Other methods (e.g.,  \cite{liu2019forecasting}) typically do not differentiate between left and right hands. See the first two rows of Table \ref{Ablations} for results.



\subsubsection{Next Active Object and Contacted Objects}
We isolate the Active and Next Active Object predictions of $o_t = \{\psi_{t_r}, \psi_{t_l}, \gamma_{t_r}, \gamma_{t_l}\}$, defined in Section \ref{sec:ego-omg}. We train the GCN stream of Ego-OMG over ablations excluding elements of $o_t$. These ablations evaluate the extent to which inclusion of Next Active Object predictions and the inclusion of Active Object Predictions benefit Ego-OMG performance. See rows 3 and 4 of Table \ref{Ablations} for results.



\begin{table}
\centering
\begin{tabular}{l*{6}c}

\toprule
Ablations & Anticipation & Prediction  & \\ 
\midrule
Ours      & \textbf{12.81}        & \textbf{15.44}           \\
Joint   & 12.12  & 14.63   \\
AO only & 10.91  & 13.96  \\
NAO only & 7.86  & 8.50  \\
\bottomrule
\end{tabular}
\vspace{1em}
\caption{Ablation experiments showing anticipation and prediction top-1 accuracy, performed over Anticipation Module components. Ours refers to non-ablated implementation; Joint collapses left vs. right hand distinctions; ``AO Only''
refers to ``Active Object Only''; ``NAO Only'' refers to 'Next Active Object Only'.
}
\label{Ablations}
\end{table}

\begin{table}
\centering

  \vspace*{.5\baselineskip}
 \begin{tabular}{ r|c|c }
\multicolumn{1}{r}{}
 &  \multicolumn{1}{c|}{GCN}
 & \multicolumn{1}{c}{No GCN} \\
\hline{}
GloVe Vectors & \textbf{12.81} & 11.79\\
\hline{}
Identity Mat. & 6.67 &  3.62\\
\end{tabular}
\caption{Action anticipation accuracies over validation set with anticipation time $\tau_a = 1$ second, over GCN and GloVe embedding ablations.}
\label{table:gcn_glove}
\vspace{-.5cm}

\end{table}

\subsection{GCN Ablations}
In evaluating the utility provided by graph embeddings effected through GCN layers, we compare two versions of Ego-OMG: One where graph nodes $V$ are embedded through a GCN, and the other where nodes $V$ are left unaltered before the node sequence observed from the video clip is fed into the LSTM.

In evaluating the utility provided by word embedding when representing states $s_i$, we compare two versions of Ego-OMG: One where the initialization of input matrix $X$ is set to features extracted from a pre-trained GloVe-600 model through methods discussed in Section \ref{subsec:construction}, and the other where $X$ is set to the identity matrix. 

Table \ref{table:gcn_glove} illustrates results over joint ablations for the two sets of comparisons. The use of graph convolutions in conjunction with GloVe embeddings outperforms ablations. 

\vspace{-.2cm}
\section{Conclusion}
\label{sec:conclusion}


We have introduced methods to produce hand/object centric representations for egocentric video which are of utility to action understanding. Contact Anticipation Maps provide time-to-contact predictions between hands and the environment, and Next Active Object Segmentations provide predictions localizing the Next Active Object. In training the Anticipation Module to produce these representations we gather contact annotations and object segmentations over a portion of the EPIC Kitchens dataset. We achieve state-of-the-art results over the EPIC Kitchens Action Anticipation Challenge - achieving \nth{1} and \nth{2} place on the unseen and seen test sets, respectively - through feeding our representations through Ego-OMG, our state of the art action anticipation and action prediction architecture. We release our predictions over the EGTEA dataset and provide ablation studies evaluating the utility of individual system characteristics.


\vspace{-.2cm}
\section{Acknowledgements}
The support of Northrop Grumman Mission Systems University Research Program, of ONR under grant award N00014-17-1-2622, and the support of the National Science Foundation under grants BCS 1824198 and CNS 1544787 is gratefully acknowledged.

\bibliographystyle{IEEEtran}
\bibliography{references}

\begin{thebibliography}{10}
\providecommand{\url}[1]{#1}
\csname url@samestyle\endcsname
\providecommand{\newblock}{\relax}
\providecommand{\bibinfo}[2]{#2}
\providecommand{\BIBentrySTDinterwordspacing}{\spaceskip=0pt\relax}
\providecommand{\BIBentryALTinterwordstretchfactor}{4}
\providecommand{\BIBentryALTinterwordspacing}{\spaceskip=\fontdimen2\font plus
\BIBentryALTinterwordstretchfactor\fontdimen3\font minus
  \fontdimen4\font\relax}
\providecommand{\BIBforeignlanguage}[2]{{%
\expandafter\ifx\csname l@#1\endcsname\relax
\typeout{** WARNING: IEEEtran.bst: No hyphenation pattern has been}%
\typeout{** loaded for the language `#1'. Using the pattern for}%
\typeout{** the default language instead.}%
\else
\language=\csname l@#1\endcsname
\fi
#2}}
\providecommand{\BIBdecl}{\relax}
\BIBdecl

\bibitem{sexton2018anticipation}
K.~Sexton, A.~Johnson, A.~Gotsch, A.~A. Hussein, L.~Cavuoto, and K.~A. Guru,
  ``Anticipation, teamwork and cognitive load: chasing efficiency during
  robot-assisted surgery,'' \emph{BMJ quality \& safety}, vol.~27, no.~2, pp.
  148--154, 2018.

\bibitem{vesper2014support}
C.~Vesper, ``How to support action prediction: Evidence from human coordination
  tasks,'' in \emph{The 23rd IEEE International Symposium on Robot and Human
  Interactive Communication}.\hskip 1em plus 0.5em minus 0.4em\relax IEEE,
  2014, pp. 655--659.

\bibitem{damen2018scaling}
D.~Damen, H.~Doughty, G.~Maria~Farinella, S.~Fidler, A.~Furnari, E.~Kazakos,
  D.~Moltisanti, J.~Munro, T.~Perrett, W.~Price \emph{et~al.}, ``Scaling
  egocentric vision: The epic-kitchens dataset,'' in \emph{Proceedings of the
  European Conference on Computer Vision (ECCV)}, 2018, pp. 720--736.

\bibitem{li2018eye}
Y.~Li, M.~Liu, and J.~M. Rehg, ``In the eye of beholder: Joint learning of gaze
  and actions in first person video,'' in \emph{Proceedings of the European
  Conference on Computer Vision (ECCV)}, 2018, pp. 619--635.

\bibitem{pirsiavash2012detecting}
H.~Pirsiavash and D.~Ramanan, ``Detecting activities of daily living in
  first-person camera views,'' in \emph{2012 IEEE conference on computer vision
  and pattern recognition}.\hskip 1em plus 0.5em minus 0.4em\relax IEEE, 2012,
  pp. 2847--2854.

\bibitem{tresilian1995perceptual}
J.~Tresilian, ``Perceptual and cognitive processes in time-to-contact
  estimation: Analysis of prediction-motion and relative judgment tasks,''
  \emph{Perception \& Psychophysics}, vol.~57, no.~2, pp. 231--245, 1995.

\bibitem{zago2009visuo}
M.~Zago, J.~McIntyre, P.~Senot, and F.~Lacquaniti, ``Visuo-motor coordination
  and internal models for object interception,'' \emph{Experimental Brain
  Research}, vol. 192, no.~4, pp. 571--604, 2009.

\bibitem{jeannerod1984timing}
M.~Jeannerod, ``The timing of natural prehension movements,'' \emph{Journal of
  motor behavior}, vol.~16, no.~3, pp. 235--254, 1984.

\bibitem{stadler2012movement}
W.~Stadler, A.~Springer, J.~Parkinson, and W.~Prinz, ``Movement kinematics
  affect action prediction: comparing human to non-human point-light actions,''
  \emph{Psychological research}, vol.~76, no.~4, pp. 395--406, 2012.

\bibitem{dessalene2020egocentric}
E.~Dessalene, M.~Maynord, C.~Devaraj, C.~Fermuller, and Y.~Aloimonos,
  ``Egocentric object manipulation graphs,'' \emph{arXiv preprint
  arXiv:2006.03201}, 2020.

\bibitem{Tran_2019_ICCV}
D.~Tran, H.~Wang, L.~Torresani, and M.~Feiszli, ``Video classification with
  channel-separated convolutional networks,'' in \emph{IEEE International
  Conference on Computer Vision (ICCV)}, October 2019.

\bibitem{carreira2017quo}
J.~Carreira and A.~Zisserman, ``Quo vadis, action recognition? a new model and
  the kinetics dataset,'' in \emph{Proceedings of the IEEE Conference on
  Computer Vision and Pattern Recognition}, 2017, pp. 6299--6308.

\bibitem{kipf2016semi}
T.~N. Kipf and M.~Welling, ``Semi-supervised classification with graph
  convolutional networks,'' \emph{arXiv preprint arXiv:1609.02907}, 2016.

\bibitem{kong2018human}
Y.~Kong and Y.~Fu, ``Human action recognition and prediction: A survey,''
  \emph{arXiv preprint arXiv:1806.11230}, 2018.

\bibitem{vondrick2015anticipating}
C.~Vondrick, H.~Pirsiavash, and A.~Torralba, ``Anticipating the future by
  watching unlabeled video.''

\bibitem{gao2018ican}
C.~Gao, Y.~Zou, and J.-B. Huang, ``ican: Instance-centric attention network for
  human-object interaction detection,'' \emph{arXiv preprint arXiv:1808.10437},
  2018.

\bibitem{tanke2019human}
J.~Tanke and J.~Gall, ``Human motion anticipation with symbolic label,''
  \emph{arXiv preprint arXiv:1912.06079}, 2019.

\bibitem{abu2018will}
Y.~Abu~Farha, A.~Richard, and J.~Gall, ``When will you do what?-anticipating
  temporal occurrences of activities,'' in \emph{Proceedings of the IEEE
  Conference on Computer Vision and Pattern Recognition}, 2018, pp. 5343--5352.

\bibitem{furnari2019would}
A.~Furnari and G.~M. Farinella, ``What would you expect? anticipating
  egocentric actions with rolling-unrolling lstms and modality attention,'' in
  \emph{Proceedings of the IEEE International Conference on Computer Vision},
  2019, pp. 6252--6261.

\bibitem{liu2019forecasting}
M.~Liu, S.~Tang, Y.~Li, and J.~Rehg, ``Forecasting human object interaction:
  Joint prediction of motor attention and egocentric activity,'' \emph{arXiv
  preprint arXiv:1911.10967}, 2019.

\bibitem{nagarajan2020ego}
T.~Nagarajan, Y.~Li, C.~Feichtenhofer, and K.~Grauman, ``Ego-topo: Environment
  affordances from egocentric video,'' \emph{arXiv preprint arXiv:2001.04583},
  2020.

\bibitem{ke2019time}
Q.~Ke, M.~Fritz, and B.~Schiele, ``Time-conditioned action anticipation in one
  shot,'' in \emph{Proceedings of the IEEE Conference on Computer Vision and
  Pattern Recognition}, 2019, pp. 9925--9934.

\bibitem{bambach2015lending}
S.~Bambach, S.~Lee, D.~J. Crandall, and C.~Yu, ``Lending a hand: Detecting
  hands and recognizing activities in complex egocentric interactions,'' in
  \emph{Proceedings of the IEEE International Conference on Computer Vision},
  2015, pp. 1949--1957.

\bibitem{urooj2018analysis}
A.~Urooj and A.~Borji, ``Analysis of hand segmentation in the wild,'' in
  \emph{Proceedings of the IEEE Conference on Computer Vision and Pattern
  Recognition}, 2018, pp. 4710--4719.

\bibitem{yang2015grasp}
Y.~Yang, C.~Fermuller, Y.~Li, and Y.~Aloimonos, ``Grasp type revisited: A
  modern perspective on a classical feature for vision,'' in \emph{Proceedings
  of the IEEE Conference on Computer Vision and Pattern Recognition}, 2015, pp.
  400--408.

\bibitem{garcia2018first}
G.~Garcia-Hernando, S.~Yuan, S.~Baek, and T.-K. Kim, ``First-person hand action
  benchmark with rgb-d videos and 3d hand pose annotations,'' in
  \emph{Proceedings of the IEEE conference on computer vision and pattern
  recognition}, 2018, pp. 409--419.

\bibitem{luvizon2017learning}
D.~C. Luvizon, H.~Tabia, and D.~Picard, ``Learning features combination for
  human action recognition from skeleton sequences,'' \emph{Pattern Recognition
  Letters}, vol.~99, pp. 13--20, 2017.

\bibitem{tekin2019h+}
B.~Tekin, F.~Bogo, and M.~Pollefeys, ``H+ o: Unified egocentric recognition of
  3d hand-object poses and interactions,'' in \emph{Proceedings of the IEEE
  Conference on Computer Vision and Pattern Recognition}, 2019, pp. 4511--4520.

\bibitem{baradel2018object}
F.~Baradel, N.~Neverova, C.~Wolf, J.~Mille, and G.~Mori, ``Object level visual
  reasoning in videos,'' in \emph{Proceedings of the European Conference on
  Computer Vision (ECCV)}, 2018, pp. 105--121.

\bibitem{li2015delving}
Y.~Li, Z.~Ye, and J.~M. Rehg, ``Delving into egocentric actions,'' in
  \emph{Proceedings of the IEEE Conference on Computer Vision and Pattern
  Recognition}, 2015, pp. 287--295.

\bibitem{furnari2017next}
A.~Furnari, S.~Battiato, K.~Grauman, and G.~M. Farinella, ``Next-active-object
  prediction from egocentric videos,'' \emph{Journal of Visual Communication
  and Image Representation}, vol.~49, pp. 401--411, 2017.

\bibitem{nagarajan2019grounded}
T.~Nagarajan, C.~Feichtenhofer, and K.~Grauman, ``Grounded human-object
  interaction hotspots from video,'' in \emph{Proceedings of the IEEE
  International Conference on Computer Vision}, 2019, pp. 8688--8697.

\bibitem{xiao2019reasoning}
T.~Xiao, Q.~Fan, D.~Gutfreund, M.~Monfort, A.~Oliva, and B.~Zhou, ``Reasoning
  about human-object interactions through dual attention networks,'' in
  \emph{Proceedings of the IEEE International Conference on Computer Vision},
  2019, pp. 3919--3928.

\bibitem{simonyan2014very}
K.~Simonyan and A.~Zisserman, ``Very deep convolutional networks for
  large-scale image recognition,'' \emph{arXiv preprint arXiv:1409.1556}, 2014.

\bibitem{jain2016structural}
A.~Jain, A.~R. Zamir, S.~Savarese, and A.~Saxena, ``Structural-rnn: Deep
  learning on spatio-temporal graphs,'' in \emph{Proceedings of the IEEE
  Conference on Computer Vision and Pattern Recognition}, 2016, pp. 5308--5317.

\bibitem{Wang_2018_ECCV}
X.~Wang and A.~Gupta, ``Videos as space-time region graphs,'' in \emph{European
  Conference on Computer Vision (ECCV)}, September 2018.

\bibitem{soran2015generating}
B.~Soran, A.~Farhadi, and L.~Shapiro, ``Generating notifications for missing
  actions: Don't forget to turn the lights off!'' in \emph{Proceedings of the
  IEEE International Conference on Computer Vision}, 2015, pp. 4669--4677.

\bibitem{lan2015action}
T.~Lan, Y.~Zhu, A.~Roshan~Zamir, and S.~Savarese, ``Action recognition by
  hierarchical mid-level action elements,'' in \emph{Proceedings of the IEEE
  international conference on computer vision}, 2015, pp. 4552--4560.

\bibitem{furnari2018leveraging}
A.~Furnari, S.~Battiato, and G.~Maria~Farinella, ``Leveraging uncertainty to
  rethink loss functions and evaluation measures for egocentric action
  anticipation,'' in \emph{Proceedings of the European Conference on Computer
  Vision (ECCV)}, 2018, pp. 389--405.

\bibitem{ren2015faster}
S.~Ren, K.~He, R.~Girshick, and J.~Sun, ``Faster r-cnn: Towards real-time
  object detection with region proposal networks,'' in \emph{Advances in neural
  information processing systems}, 2015, pp. 91--99.

\bibitem{perez2013tv}
J.~S. P{\'e}rez, E.~Meinhardt-Llopis, and G.~Facciolo, ``Tv-l1 optical flow
  estimation,'' \emph{Image Processing On Line}, vol. 2013, pp. 137--150, 2013.

\bibitem{ronneberger2015u}
O.~Ronneberger, P.~Fischer, and T.~Brox, ``U-net: Convolutional networks for
  biomedical image segmentation,'' in \emph{International Conference on Medical
  image computing and computer-assisted intervention}.\hskip 1em plus 0.5em
  minus 0.4em\relax Springer, 2015, pp. 234--241.

\bibitem{chollet2017xception}
F.~Chollet, ``Xception: Deep learning with depthwise separable convolutions,''
  in \emph{Proceedings of the IEEE Conference on Computer Vision and Pattern
  Recognition}, 2017, pp. 1251--1258.

\bibitem{ghadiyaram2019large}
D.~Ghadiyaram, D.~Tran, and D.~Mahajan, ``Large-scale weakly-supervised
  pre-training for video action recognition,'' in \emph{Proceedings of the IEEE
  Conference on Computer Vision and Pattern Recognition}, 2019, pp.
  12\,046--12\,055.

\bibitem{hochreiter1997long}
S.~Hochreiter and J.~Schmidhuber, ``Long short-term memory,'' \emph{Neural
  Computation}, vol.~9, no.~8, pp. 1735--1780, 1997.

\bibitem{pennington-etal-2014-glove}
\BIBentryALTinterwordspacing
J.~Pennington, R.~Socher, and C.~Manning, ``{G}love: Global vectors for word
  representation,'' in \emph{Proceedings of the 2014 Conference on Empirical
  Methods in Natural Language Processing ({EMNLP})}.\hskip 1em plus 0.5em minus
  0.4em\relax Doha, Qatar: Association for Computational Linguistics, Oct.
  2014, pp. 1532--1543. [Online]. Available:
  \url{https://www.aclweb.org/anthology/D14-1162}
\BIBentrySTDinterwordspacing

\bibitem{shou2016temporal}
Z.~Shou, D.~Wang, and S.-F. Chang, ``Temporal action localization in untrimmed
  videos via multi-stage cnns,'' in \emph{Proceedings of the IEEE Conference on
  Computer Vision and Pattern Recognition}, 2016, pp. 1049--1058.

\bibitem{wang2016temporal}
L.~Wang, Y.~Xiong, Z.~Wang, Y.~Qiao, D.~Lin, X.~Tang, and L.~Van~Gool,
  ``Temporal segment networks: Towards good practices for deep action
  recognition,'' in \emph{European Conference on Computer Vision}.\hskip 1em
  plus 0.5em minus 0.4em\relax Springer, 2016, pp. 20--36.

\bibitem{camporese2020knowledge}
G.~Camporese, P.~Coscia, A.~Furnari, G.~M. Farinella, and L.~Ballan,
  ``Knowledge distillation for action anticipation via label smoothing,''
  \emph{arXiv preprint arXiv:2004.07711}, 2020.

\bibitem{selvaraju2017grad}
R.~R. Selvaraju, M.~Cogswell, A.~Das, R.~Vedantam, D.~Parikh, and D.~Batra,
  ``Grad-cam: Visual explanations from deep networks via gradient-based
  localization,'' in \emph{Proceedings of the IEEE international conference on
  computer vision}, 2017, pp. 618--626.

\bibitem{Kummerer_2017_ICCV}
M.~Kummerer, T.~S.~A. Wallis, L.~A. Gatys, and M.~Bethge, ``Understanding low-
  and high-level contributions to fixation prediction,'' in \emph{The IEEE
  International Conference on Computer Vision (ICCV)}, Oct 2017.

\bibitem{bewley2016simple}
A.~Bewley, Z.~Ge, L.~Ott, F.~Ramos, and B.~Upcroft, ``Simple online and
  realtime tracking,'' in \emph{2016 IEEE International Conference on Image
  Processing (ICIP)}.\hskip 1em plus 0.5em minus 0.4em\relax IEEE, 2016, pp.
  3464--3468.

\end{thebibliography}

\vskip -2\baselineskip plus -1fil

\begin{IEEEbiography}[{\includegraphics[width=1in,height=1.25in,clip]{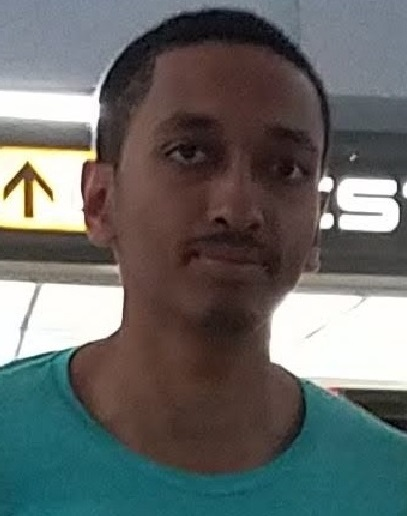}}]{Eadom Dessalene}
is currently doing his PhD in the department of Computer Science at the University of Maryland College Park. He is advised by Yiannis Aloimonos and Cornelia Fermuller. His research interests include computer vision and reinforcement learning. His recent work has focused on developing structured representations of video for downstream applications in AI and robotics. 
\end{IEEEbiography}
\vskip -3\baselineskip plus -1fil
\begin{IEEEbiography}[{\includegraphics[width=1in,height=1.25in,clip]{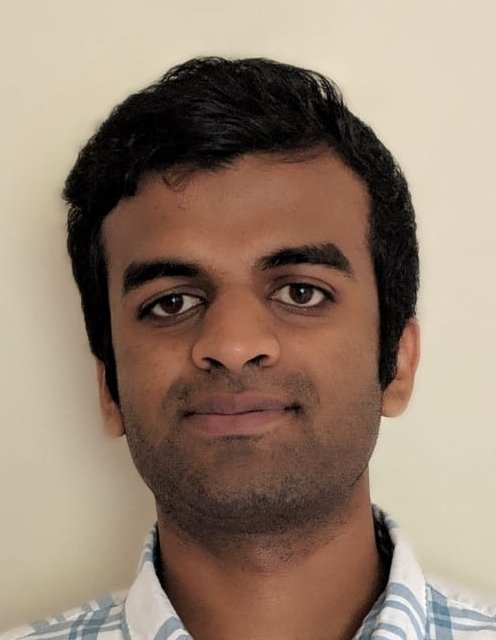}}]{Chinmaya Devaraj}
is currently doing his Ph.D. in Electrical and Computer  Engineering at the University of Maryland College Park. He is advised by Prof Yiannis Aloimonos and Dr. Cornelia Fermuller. His Ph.D. thesis is on action understanding. Prior to this, he graduated with a B.Tech In Electrical and Electronic Engineering from the National Institute of Technology Karnataka, Surathkal, India.
\end{IEEEbiography}
\vskip -3\baselineskip plus -1fil
\begin{IEEEbiography}[{\includegraphics[width=1in,height=1.25in,clip]{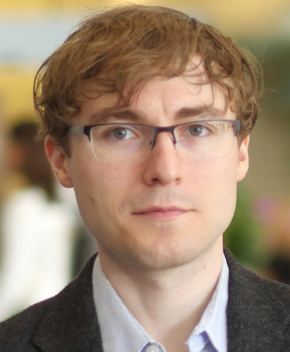}}]{Michael Maynord} is a PhD candidate in the department of Computer Science at the University of Maryland College Park, advised by Yiannis Aloimonos and Cornelia Fermuller. His background encompasses symbolic Artificial Intelligence, including cognitive architectures, Computer Vision, including action understanding, and methods integrating AI and CV.
\end{IEEEbiography}
\vskip -3\baselineskip plus -1fil
\begin{IEEEbiography}[{\includegraphics[width=1in,height=1.25in,clip]{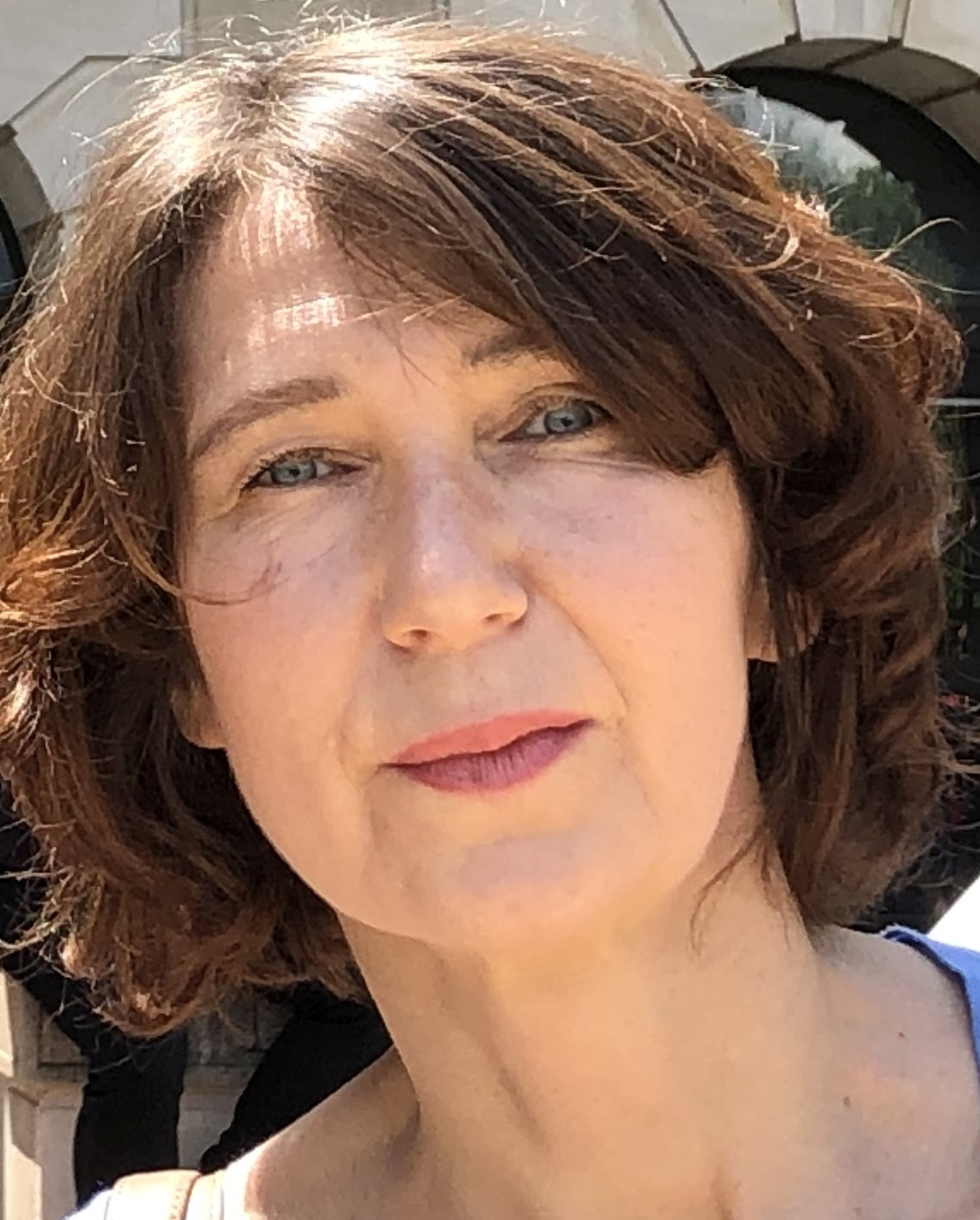}}]{Cornelia Ferm\"uller}
is a Research Scientist at the University of Maryland Institute for Advanced Computer Studies. She holds a Ph.D. from the Vienna University of Technology, Austria (1993) and an M.S. from the Graz University of Technology (1989), both in Applied Mathematics. Her research interest has been to understand principles of active vision systems and develop biological-inspired methods, especially in the area of motion. Her recent work has focused on human action interpretation and the development of event-based motion algorithms.  
\end{IEEEbiography}
\vskip -3\baselineskip plus -1fil
\begin{IEEEbiography}[{\includegraphics[width=1in,height=1.25in,clip]{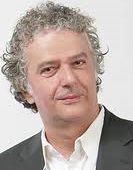}}]{Yiannis Aloimonos}
is Professor of Computational Vision and Intelligence at the Department of Computer Science, University of Maryland, College Park, and the Director of the Computer Vision Laboratory at the Institute for Advanced Computer Studies (UMIACS). He is interested in Active Perception and the modeling of vision as an active, dynamic process for real time robotic systems. For the past five years he has been working on bridging signals and symbols, specifically on the relationship of vision to reasoning, action and language.
\end{IEEEbiography}

\newpage

\ifCLASSOPTIONcaptionsoff
  \newpage
\fi

%
\end{document}


%
\title{Forecasting Action through Contact Representations from First Person Video}
\author{Eadom Dessalene\textbf{*}\thanks{$\bullet$ \:The authors are with the department of Computer Science, University of Maryland, College Park, MD, 20742}\thanks{$\bullet$ \:*Equal contribution}, Chinmaya Devaraj\textbf{*}, Michael Maynord\textbf{*}, Cornelia Ferm\"uller,
and Yiannis Aloimonos}

\maketitle

\IEEEpeerreviewmaketitle


%

\section{Supplementary Material}

\subsection{Datasets}

\subsubsection{Collection}
\begin{figure}[!t]
\includegraphics[width=0.48\textwidth]{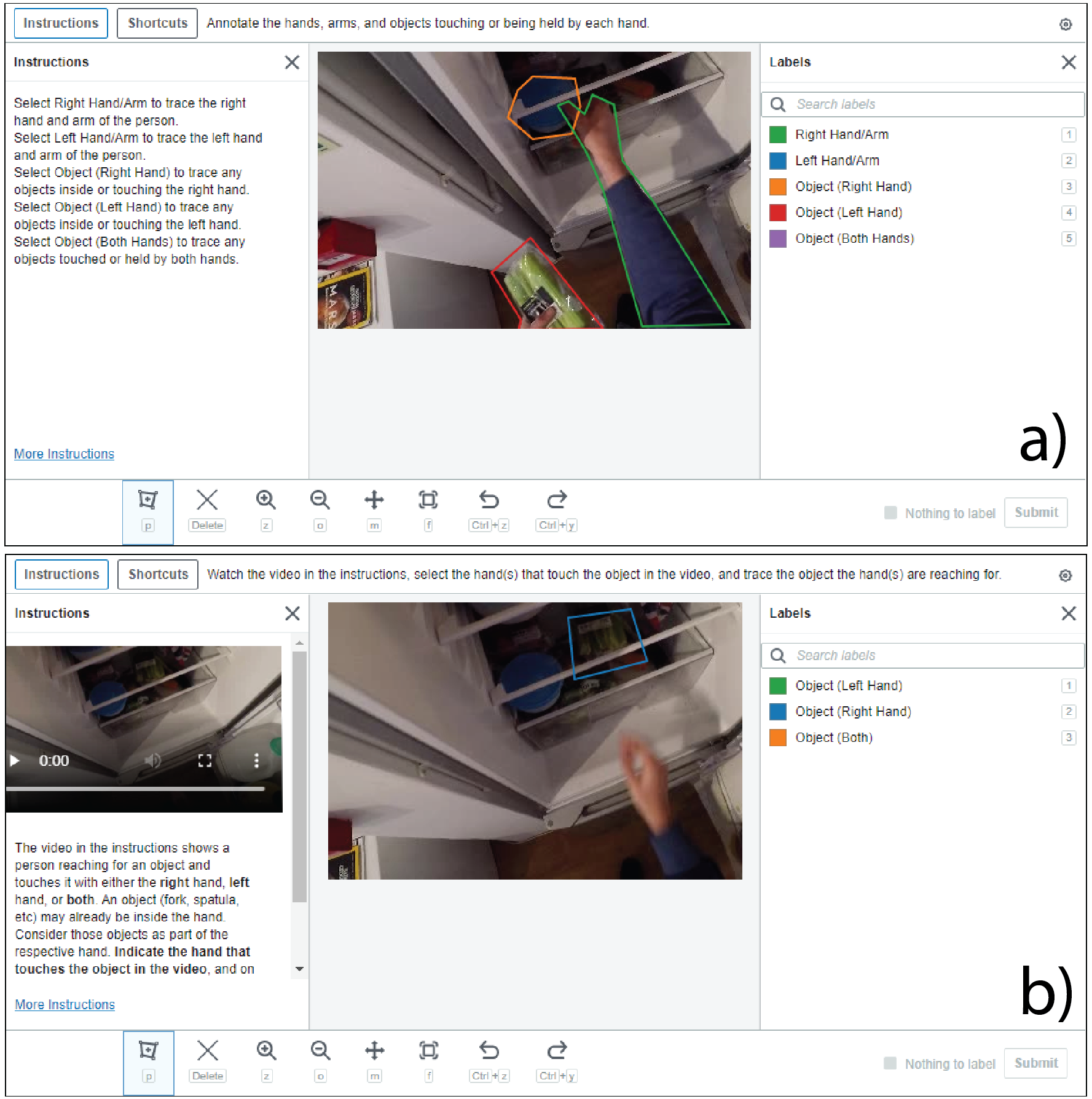}
  \caption{Our interfaces for collection of annotations over the Amazon Mechanical Turk platform, where \textbf{(a)} is the interface for the collection of the body/contacted objects, and \textbf{(b)} is the interface for the collection of the Next Active Object annotations. In \textbf{(a)}, workers are instructed to annotate all instances of hands and objects in contact by selecting the appropriate label in the right panel and tracing the body of the hand or object involved. In \textbf{(b)}, workers are instructed to first view the full video of the point to point hand movement, as the Next Active Object annotation relies on information only available in the future. They then select the side(s) involved in the reaching movement in the right panel and trace the body of the object involved.
  }
  \label{fig:interface}
\end{figure}

As discussed in Section 3.1.1, we collect our dataset by organizing clips that correspond to point-to-point hand movements, where the hand(s) involved and the Next Active Object are visible. The clips are croudsourced onto the Amazon Mechanical Turk platform, where $103K$ responses were collected from $162$ workers, resulting in $45K$ final responses after inspecting annotations of each worker and preventing workers whose jobs did not meet a satisfactory level of performance from submitting further responses. Workers are compensated $\$0.03$ for each submitted response. The annotations for the body and objects in contact are croudsourced separately from the annotations for the Next Active Object, with the interface for both shown in Figure \ref{fig:interface}. In the end, a single annotation per sample is collected and used for the training of the Anticipation Module.

\subsubsection{EPIC Kitchens}
The EPIC Kitchens dataset \cite{damen2018scaling} is a large egocentric video dataset, captured by $32$ subjects in $32$ different kitchens. The videos consist of daily kitchen activities where participants were simply asked to record their interactions in their native kitchen environments (no instructional scripts were given to the subjects). Each video contains several annotated action segments, each associated a $(verb, noun)$ action label, where there are $125$ unique verbs and $352$ unique nouns, where $2,513$ unique actions are present in the dataset. In addition, there are $400K$ bounding box annotations of objects over the entire dataset. We conduct all quantitative experiments over this dataset.

\subsubsection{EGTEA Gaze+}
The EGTEA Gaze+ dataset \cite{li2018eye} is another egocentric video dataset, captured by $32$ subjects, where participants are given instructions to prepare meals. To demonstrate the generalization capabilities of the Anticipation Module to other egocentric datasets, we provide the outputs of the Anticipation Module pre-trained on EPIC Kitchens over the entirety of the EGTEA Gaze+ dataset. A Google Drive link is provided\footnote{ \vspace{-5mm} Google Drive link at : https://drive.google.com/drive/folders/1AIZ93d37g0mJaHclANhXYVyp2jFQtfCS?usp=sharing}

\subsection{EPIC Kitchens Challenge}
We demonstrate state-of-the-art performance over the recent EPIC Kitchens Action Anticipation Challenge, achieving 1st place on the EPIC Kitchens Action Anticipation Challenge unseen test set, and 2nd place on the seen test set, and outperform all previously published approaches without any use of ensembling, unlike many competing approaches. Below we describe the action anticipation results reported in Table 1.

\begin{itemize}
	\item \textbf{2SCNN} \cite{shou2016temporal} is a two-stream CNN training 2D ConvNets on separate RGB and flow streams before performing a late-fusion.
	\item \textbf{TSN (RGB)} \cite{wang2016temporal} is a temporal segment network trained on a single stream of RGB video.	
	\item \textbf{TSN + MCE} \cite{furnari2018leveraging} is a temporal segment network trained with a novel loss formulation to address the inherent ambiguity in action anticipation.
	\item \textbf{RULSTM} \cite{furnari2019would} is a state of the art architecture with two separate LSTMs and an attention formulation applied over features obtained from RGB, flow, and object detections.
	\item \textbf{Camp et al.} \cite{camporese2020knowledge} is a loss formulation representing verbs and objects with pre-computed word embeddings to overcome the overwhelming number of unique actions in EPIC Kitchens.
	\item \textbf{Liu et al.} \cite{liu2019forecasting} is a joint architecture that relies on additional supervisory signals to predict motor attention and interaction hotspots before performing action anticipation. This is the closest work to our approach.
	\item \textbf{Ours} The combined Anticipation Module and Ego-OMG.
\end{itemize}

\begin{figure}[H]
\includegraphics[width=8cm]{Experimental Setup 3.png}
  \caption{Jeannerod put forth the hypothesis that velocity profiles of reaching actions follow a well formed bell-shaped distribution \cite{jeannerod1984timing}. We sample multiple hand trajectories involved in reaching motions in a lab setting using a Vicon motion tracking system. The three curves represent the velocity profiles of three separate trajectory distances at table positions A, B and C. Solid lines indicate mean velocity values across multiple runs; the upper and lower bounds of the shaded regions lie one standard deviation from the mean.
  }
  \label{fig:bell}
\end{figure}

\subsection{Reaching Experiments}
To illustrate the velocity profile of reaching in hand manipulation actions we gathered position information through a Vicon motion capture system of simple reaching actions. The Vicon system gives sub-millimeter precision 3-dimensional positional information of tracked objects. Figure \ref{fig:bell} illustrates a typical velocity profile of a reaching action. Our observations of velocity profiles are consistent with Jeannerod's hypothesis \cite{jeannerod1984timing}.

\subsection{Example Graph Representation}
\begin{figure}[H]
\includegraphics[width=0.48\textwidth]{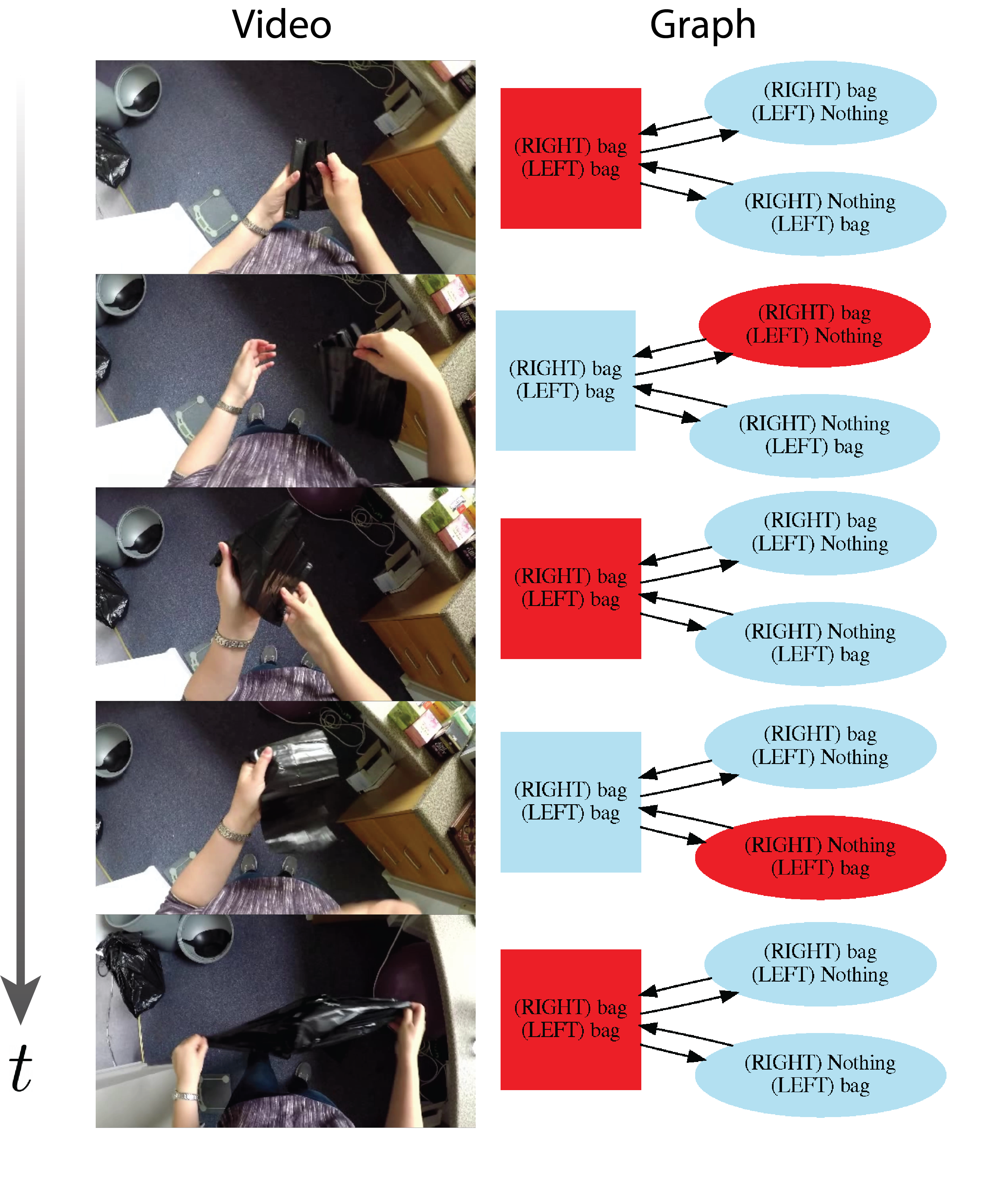}
  \centering
  \caption{An example graph representation capturing the activity of opening a trash bag over the span of $35$ seconds. For ease of visualization we exclude Next Active Object predictions and display only Active Object predictions for each hand. By representing videos as structured sequences of contact-based hand-object interactions, our method captures the temporal semantics underlying longer activity in video, whereas many previous end-to-end video architectures are only capable of modelling videos segments on the order of $1 - 2$ seconds long.}
 \label{fig:graph_representation}
\end{figure}

Figure \ref{fig:graph_representation} illustrates an example graph derived from a 30 second EPIC Kitchens clip involving the opening of a trash bag, illustrating the semantic relations which Ego-OMG models. For easier illustration we include in states only the object in direct contact with each hand (full states include objects involved in anticipated contact).

\bibliographystyle{IEEEtran}
\bibliography{references}

%



